\documentclass[preprint,12pt]{elsarticle}

\usepackage{graphicx}%
\usepackage{multirow}%
\usepackage{amsmath,amssymb,amsfonts}%
\usepackage{amsthm}%
\usepackage{mathrsfs}%
\usepackage[title]{appendix}%
\usepackage{xcolor}%
\usepackage{textcomp}%
\usepackage{manyfoot}%
\usepackage{booktabs}%
\usepackage{algorithmic}
\usepackage{algorithm}
\usepackage{listings}%
\usepackage{bm}
\usepackage{array}
\usepackage[caption=false,font=normalsize,labelfont=sf,textfont=sf]{subfig}
\usepackage{stfloats}
\usepackage{url}
\usepackage{verbatim}
\usepackage{times}
\usepackage{epsfig}
\usepackage{setspace}
\usepackage{bbding}
\usepackage{makecell,multirow,diagbox}
\usepackage{bm}
\usepackage[accsupp]{axessibility}
\newcommand{\eg}{e.g.} %\@\xspace
\newcommand{\ie}{i.e.}  %\@\xspace
\usepackage[accsupp]{axessibility}  % Improves PDF readability for those with disabilities.
\journal{Information Sciences}

\begin{document}

\begin{frontmatter}



\title{Aggregating {Nearest} Sharp Features via Hybrid Transformers for Video Deblurring}


%\author{} %% Author name
%
%%% Author affiliation
%\affiliation{organization={},%Department and Organization
\author[1,2]{Wei Shang} %\email{rendongweihit@gmail.com}

\author[1]{Dongwei Ren\corref{cor1}}
% \author*[1,2]{\fnm{Wei} \sur{Shang}}\email{csweishang@gmail.com}

\author[1]{Yi Yang} %\email{22s003027@stu.hit.edu.cn}

\author[1]{Wangmeng Zuo}  %\email{wmzuo@hit.edu.cn}

%\affil[1]{\orgdiv{School of Computer Science and Technology}, \orgname{Harbin Institute of Technology}, \orgaddress{\postcode{150001}, \country{China}}}
%
%\affil[2]{\orgdiv{Computer Science}, \orgname{City University of Hong Kong}, \orgaddress{\postcode{999077}, \country{Hong Kong}}}

\affiliation[1]{organization={School of Computer Science and Technology, Harbin Institute of Technology},%Department and Organization
%	            addressline={}, 
	            city={Harbin},
	            postcode={150001}, 
%	            state={},
	            country={China}}
\affiliation[2]{organization={Computer Science, City University of Hong Kong},%Department and Organization
	%	            addressline={}, 
%	city={Harbin},
	postcode={999077}, 
	%	            state={},
	country={Hong Kong}}

\cortext[cor1]{Corresponding author: csdren@hit.edu.cn.}
%% Abstract
\begin{abstract}
Video deblurring methods, aiming at recovering consecutive sharp frames from a given blurry video, usually assume that the input video suffers from consecutively blurry frames. 
{However, in real-world scenarios captured by modern imaging devices, sharp frames often interspersed within the video, providing temporally nearest sharp features that can aid in the restoration of blurry frames.}
In this work, we propose a video deblurring method that leverages both neighboring frames and {existing} sharp frames using hybrid Transformers for feature aggregation.
Specifically, we first train a blur-aware detector to distinguish between sharp and blurry frames.
Then, a window-based local Transformer is employed for exploiting features from neighboring frames, where cross attention is beneficial for aggregating features from neighboring frames without explicit spatial alignment.
To aggregate {nearest} sharp features from detected sharp frames, we utilize a global Transformer with multi-scale matching capability. 
Moreover, our method can easily be extended to event-driven video deblurring by incorporating an event fusion module into the global Transformer.
% in the multi-scale scheme. 
%
Extensive experiments on benchmark datasets demonstrate that our proposed method outperforms state-of-the-art video deblurring methods as well as event-driven video deblurring methods in terms of quantitative metrics and visual quality. 
The source code and trained models are available at \url{https://github.com/shangwei5/STGTN}.
\end{abstract}



%% Keywords

\end{frontmatter}



\begin{figure}[!t]\footnotesize
	\centering
	\begin{tabular}{cccccc}
		\includegraphics[width=0.95\linewidth]{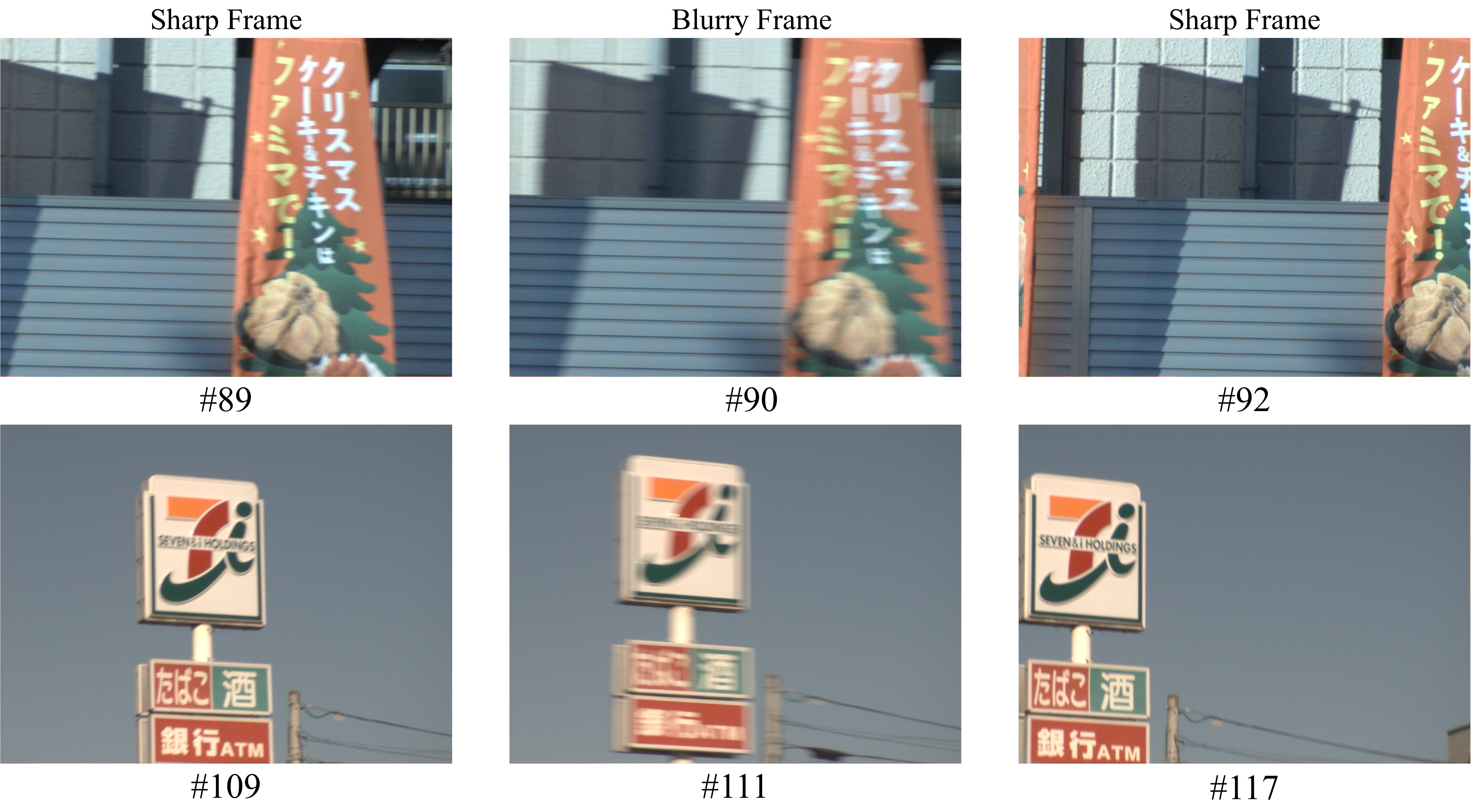}\\
	\end{tabular}
	 	\vspace{-1em}
	\caption{Examples of sharp frames observed in real-world blurry videos~\citep{zhong2023real}, where \# denotes the frame number in a captured video.
		{Nearest} sharp features extracted from these sharp frames can be leveraged to enhance the restoration of the corresponding blurry frame. }
	%Frames from the left to right are captured with increasing exposure time
	\label{fig:real_example}
	% \vspace{-1.5em}
\end{figure}

\section{Introduction} \label{sec:introduction}
With the widespread use of modern hand-held or onboard imaging devices, videos have emerged as the predominant form of media data. 
{Blur, a common artifact in video capture due to camera movement or moving objects, not only degrades visual quality but also impedes performance in downstream applications such as Simultaneous Localization and Mapping (SLAM), 3D reconstruction, and tracking. The field of video deblurring, which endeavors to recover a sequence of sharp frames from blurred frames, has garnered significant research interest in recent years~\cite{pan2020cascaded, shang2021bringing, nah2019recurrent, liang2024vrt}.}

Unlike single image deblurring~\citep{tao2018scale,ren2020neural}, video deblurring methods~\citep{tao2018scale,ren2020neural,cho2021rethinking,hyun2017online, kim2018spatio, pan2020cascaded, shang2021bringing} can leverage temporal information from adjacent blurry frames. 
%
%In literature~\citep{delbracio2015hand,tao2018scale,pan2023cascaded,liang2024vrt}, existing video deblurring methods mainly focus on the feature aggregation of adjacent blurry frames. 
%
Due to the remarkable achievements of deep learning, extensive research has been conducted on video deblurring methods, demonstrating promising performance~\citep{tao2018scale,ren2020neural,cho2021rethinking,hyun2017online, kim2018spatio, pan2020cascaded, shang2021bringing}. 
Among these methods, convolutional neural network (CNN)-based~\citep{pan2020cascaded,chen2018reblur,shang2023joint}, recurrent neural network (RNN)-based~\citep{tao2018scale,hyun2017online,nah2019recurrent} and Transformer-based~\citep{liang2024vrt,pan2023cascaded} network architectures have been extensively investigated. 
Correspondingly, the strategies for feature aggregation of adjacent blurry frames can be categorized from three perspectives, \ie, explicit feature alignment and fusion based on optical flow~\citep{pan2023cascaded,shang2021bringing}, temporal features propagation using recurrent states~\citep{nah2019recurrent}, and attention-based long-range feature dependency~\citep{liang2024vrt}. 

{Despite significant improvements in deblurring performance, most existing video deblurring networks are trained on synthetic datasets, which often leads to suboptimal generalization when applied to real-world blurry videos. A critical oversight in these methods is their failure to account for the varying degrees of blur present in video frames. }
For instance, different exposure times or varying motion speeds can result in varying degrees of blur across video frames in the real-world.
As depicted in Fig. \ref{fig:real_example}, some frames in a blurry video are extremely sharp and clean, where temporal {nearest} sharp features are available from these sharp frames. 
This phenomenon arises from varying relative motion speed or the dynamic variation of exposure phases when the auto-exposure function is activated~\citep{shang2023joint,Zhang2019video,kim2021event}, which leads to varying amounts of blur caused by accidental camera shake or moving objects {within the same video sequence}.
Some relevant works have found this phenomenon~\citep{ren_deblur, shang2021bringing}. 
In~\cite{ren_deblur}, one deblurring model is specifically trained for a given testing video by utilizing sharp frames presented in this video as pseudo ground-truth.
In~\cite{shang2021bringing}, detected sharp frames are explicitly aligned to the current blurry frame, and the features are simply concatenated and fused in a two-stage CNN framework. 
Although Transformer-based model, \eg, VRT~\citep{liang2024vrt}, is able to exploit temporal dependency of adjacent frames, {nearest} sharp features cannot be fully utilized, and the training and testing of VRT require a lot of resources. 

In this work, we propose a video deblurring method by aggregating {nearest} sharp features from detected sharp frames as well as features from neighboring frames. 
First, a blur-aware detector is trained for distinguishing sharp frames from {the input video sequence}. 
In particular, we train a blur-aware detector using a bidirectional LSTM as the backbone ~\citep{shang2021bringing}, as shown in Fig. \ref{fig:detector}. 
This detector enables us to identify two sharp frames for each blurry frame in both forward and backward directions. 
Besides cross-entropy classification loss, we introduce contrastive learning loss to improve the discrimination between sharp and blurry frames, thereby enhancing the generalization ability of our method for real-world blurry videos.
Then, hybrid Transformers are adopted for aggregating features from neighboring frames and detected sharp frames.
As shown in Fig. \ref{fig:restor}, a window-based local Transformer is adopted for aggregating features from neighboring frames, where the features of neighboring frames are fused with the current frame by cross-attention with shifted windows. 
As for detected sharp frames, a global Transformer is employed to aggregate similar {nearest} sharp features in a multi-scale scheme. 
To reduce the computational cost, global attention matrix is computed at the coarsest level.   
Moreover, we introduce an event fusion module into the global Transformer, thereby facilitating the extension of our method to event-based video deblurring and effectively bridging the gap between conventional video deblurring and event-driven video deblurring. 

Extensive experiments have been conducted on two synthetic datasets including the GOPRO dataset~\citep{nah2017deep} and REDS dataset~\citep{nah2017deep}, as well as one real-world BSD dataset~\citep{zhong2023real} to evaluate the generalization ability of video deblurring methods on real-world blurry videos.
For event-driven video deblurring, the CED dataset~\citep{scheerlinck2019ced} and RBE dataset~\citep{pan2019bringing} are employed to assess the performance of state-of-the-art methods.  
By aggregating {nearest} sharp features, our proposed method demonstrates significant quantitative improvements over existing video deblurring methods, and there is a notable enhancement in visual quality. 
Our approach also exhibits higher efficiency compared to Transformer-based methods such as VRT~\citep{liang2024vrt} and CDVD-TSPNL~\citep{pan2023cascaded}.  
Moreover, for event-driven video deblurring tasks, our event fusion module contributes significantly for improving performance without introducing substantial computational overhead.

The contributions of this work are three-fold:
\begin{itemize}
	%	\vspace{-0.07in}
	\item 
	We propose a novel video deblurring framework that utilizes hybrid Transformers to aggregate features from both detected sharp frames and neighboring frames.  
	%	\vspace{-0.07in}
	\item 
	A blur-aware detector is trained to distinguish between sharp and blurry frames, allowing us to extract {nearest} sharp features from detected sharp frames and facilitate the restoration of blurry ones.  
	
	\item 
	We introduce an event fusion module that extends our method to event-driven video deblurring. 
	Extensive experiments on synthetic and real-world blurry datasets have been conducted to demonstrate the effectiveness of our approach in video deblurring as well as event-driven video deblurring.
	
\end{itemize}

The remainder of this paper is organized as follows. 
Section \ref{sec:related} briefly reviews relevant works, Section \ref{sec:method} introduces our proposed method for video deblurring as well as extension to event-driven video deblurring, 
Section \ref{sec:experiments} presents the experimental results and ablation studies, 
and Section \ref{sec:conclusion} ends this paper with concluding remarks.

\section{Related Work} \label{sec:related}
In this section, we briefly survey relevant works including image and video deblurring, event-driven video deblurring, and Transformer-based image and video restoration methods. 

% \IEEEpubidadjcol

\subsection{Image and Video Deblurring}
For single image deblurring, deep learning-based methods~\citep{ren2020neural,tao2018scale,aittala2018burst,zhang2019deep} have been widely studied. 
\cite{tao2018scale} proposed a scale-recurrent network in a coarse-to-fine scheme to extract multi-scale features from blurry image. 
\cite{zhang2019deep} presented a deep hierarchical multi-patch network inspired by spatial pyramid matching to handle blurry images.
\cite{ren2020neural} adopted an asymmetric autoencoder and a fully-connected network to tackle image deblurring in a self-supervised manner. 
\cite{cho2021rethinking} reevaluated coarse-to-fine approach in single image deblurring and propose a fast and accurate deblurring network.

For video deblurring, \cite{hyun2017online} developed a spatial-temporal recurrent network with a dynamic temporal blending layer for latent frame restoration.
To better leverage spatial and temporal information, \cite{kim2018spatio} introduced an optical flow estimation step for aligning and aggregating information across the neighboring frames to restore latent clean frames. 
{\cite{zhang2018dynamic} used a spatially variant RNN integrated with CNNs to effectively handle the spatially variant blur in dynamic scenes. \cite{zhang2020recursive} combined the non-local block, recursive block, and temporal loss function to better capture the complex spatio-temporal patterns.}
\cite{wang2019edvr} developed deformable convolution in a pyramid manner to implicitly align adjacent frames for better leveraging temporal information.   
\cite{pan2020cascaded} proposed to simultaneously estimate the optical flow and latent frames for video deblurring with the help of temporal sharpness prior. 
The estimated optical flow from intermediate latent frames, representing motion blur information, is fed back to the reconstruction network to generate final sharp frames.
{\cite{wang2021video} utilized a spatiotemporal pyramid module to capture spatial and temporal information at multiple scales for producing more realistic sharp video frames. \cite{zhang2021multi} incorporated temporal-spatial and channel attention mechanisms to effectively model the complex blur patterns in video.}
\cite{zhong2023real} incorporated residual dense blocks into RNN cells to efficiently extract spatial features of the current frame. Additionally, a global spatio-temporal attention module is proposed to fuse effective hierarchical features from past and future frames, aiding in better deblurring of the current frame.

Existing video deblurring methods assume consecutively blurry frames, which is commonly inconsistent with real-world blurry videos.
\cite{shang2021bringing} found that some frames in a video with motion blur are sharp, and proposed to detect sharp frames in a video and then restore the current frame by the guidance of sharp frames. 
But they simply concatenate warped sharp frames and current frames for deblurring, which limits the restoration due to insufficient exploitation of sharp textures.
In this work, we propose a new framework to better exploit sharp frames for video deblurring, where {nearest} sharp features can be aggregated by hybrid Transformers.

\subsection{Event-driven Video Deblurring}

Event cameras~\citep{patrick2008128x,brandli2014240} are innovative sensors that record intensity changes in a scene at a microsecond level with slight power consumption, and have potential applications in a variety of computer vision tasks, \eg, visual tracking~\citep{mitrokhin2018event}, stereo vision~\citep{andreopoulos2018low} and optical flow estimation~\citep{liu2018adaptive}. 
A related research area focuses on utilizing the pure events to reconstruct high frame rate image sequences~\citep{rebecq2019high}. 
Recently, \cite{pan2019bringing} formulated event-driven motion deblurring as a double integral model. Yet, the noisy hard sampling mechanism of event cameras often introduces strong accumulated noise and loss of scene details. \cite{jiang2020learning} proposed a sequential formulation of event-based motion deblurring, then unfolded its optimization steps as an end-to-end deep deblurring architecture.
\cite{wang2020event} proposed an explainable network, an event-enhanced sparse learning network (eSL-Net), to recover high-quality images from event cameras.
\cite{zhang2023event} proposed a multi-patch network approach to address the limitations of contemporary deep learning multi-scale deblurring models, such as non-uniform blur, limited motion information, and robustness to spatial transformations and noise. 
\cite{yu2022learning} presented a dual sparse learning scheme by assuming the sparsity over latent images and events, and then built a deep neural network,to take into account event noises.
In this work, we introduce an event fusion module to our method, bridging the gap between video deblurring and event-driven video deblurring. 
%To utilize event data, we propose to fuse them with image data at multiple scales so as to generate images with fine texture details and well-preserved structure. 

\subsection{Transformer-based Image and Video Restoration}
Recently, Transformer-based networks have shown significant performance gains in natural language and high-level
vision tasks~\citep{dosovitskiy2020image,carion2020end}. As Transformer is optimized for effective representation learning and captures global interactions between contexts, it has shown promising performance in
several low-level vision problems~\citep{chen2021pre,liang2021swinir,zamir2021restormer,wang2021uformer}.
\cite{chen2021pre} developed a new pre-trained model (IPT) to maximally excavate the capability of Transformer for studying the low-level computer vision task. 
\cite{liang2021swinir} proposed SwinIR, a strong baseline model for image restoration based on the Swin Transformer~\citep{liu2021Swin} and validated its effectiveness across different low-level tasks.
\cite{yang2020learning} proposed to use attention mechanisms to transfer high-resolution textures from reference images to low resolution images. Low resolution and reference images are formulated as queries and keys in a Transformer, respectively. Such a design encourages joint feature learning across low resolution images and reference images, enabling the discovery of deep feature correspondences through attention and accurate texture feature transfer.
% in which deep feature correspondences can be discovered by attention, and thus accurate texture features can be transferred.
%
\cite{zamir2021restormer} proposed an efficient Transformer model, \ie, Restormer, which incorporates key designs in the building blocks, including multi-head attention and feed-forward network, to capture long-range pixel interactions while still remaining applicable to large images.
\cite{wang2021uformer} combined a hierarchical encoder-decoder network with Transformer block forming Uformer, an effective and efficient image restoration model. 
\cite{liang2024vrt} proposed a large model for video restoration with parallel frame prediction and long-range temporal dependency modeling abilities.
{\cite{cao2022vdtr} exploited the superior long-range and relation modeling capabilities of Transformer for both spatial and temporal modeling.}
In this work, we propose a new strategy for fusing neighboring frames and {nearest} sharp frames. We utilize a window-based local Transformer to fuse neighboring frames without explicit spatial alignment using optical flow. 
Additionally, a global Transformer is adopted for transferring sharp features from detected sharp frames, resulting in significant improvements for deblurring.
% A matching \& selection module is further implemented for searching sharp textures from NSFs and then fed into decoder to guide current frame deblur.

\section{Proposed Method}\label{sec:method}
In this section, we first present our proposed video deblurring framework by aggregating {nearest} sharp features, 
then the key components including blur-aware detector and hybrid Transformers are elaborated in detail, and finally our method is extended to event-driven video deblurring by introducing an event fusion module. 
\begin{figure}[!t]\footnotesize
	\centering
	%\fbox{\rule{0pt}{3in} \rule{0.9\linewidth}{0pt}}
	\begin{tabular}{lcccccc}
		\ \ \
		\includegraphics[width=0.5\linewidth]{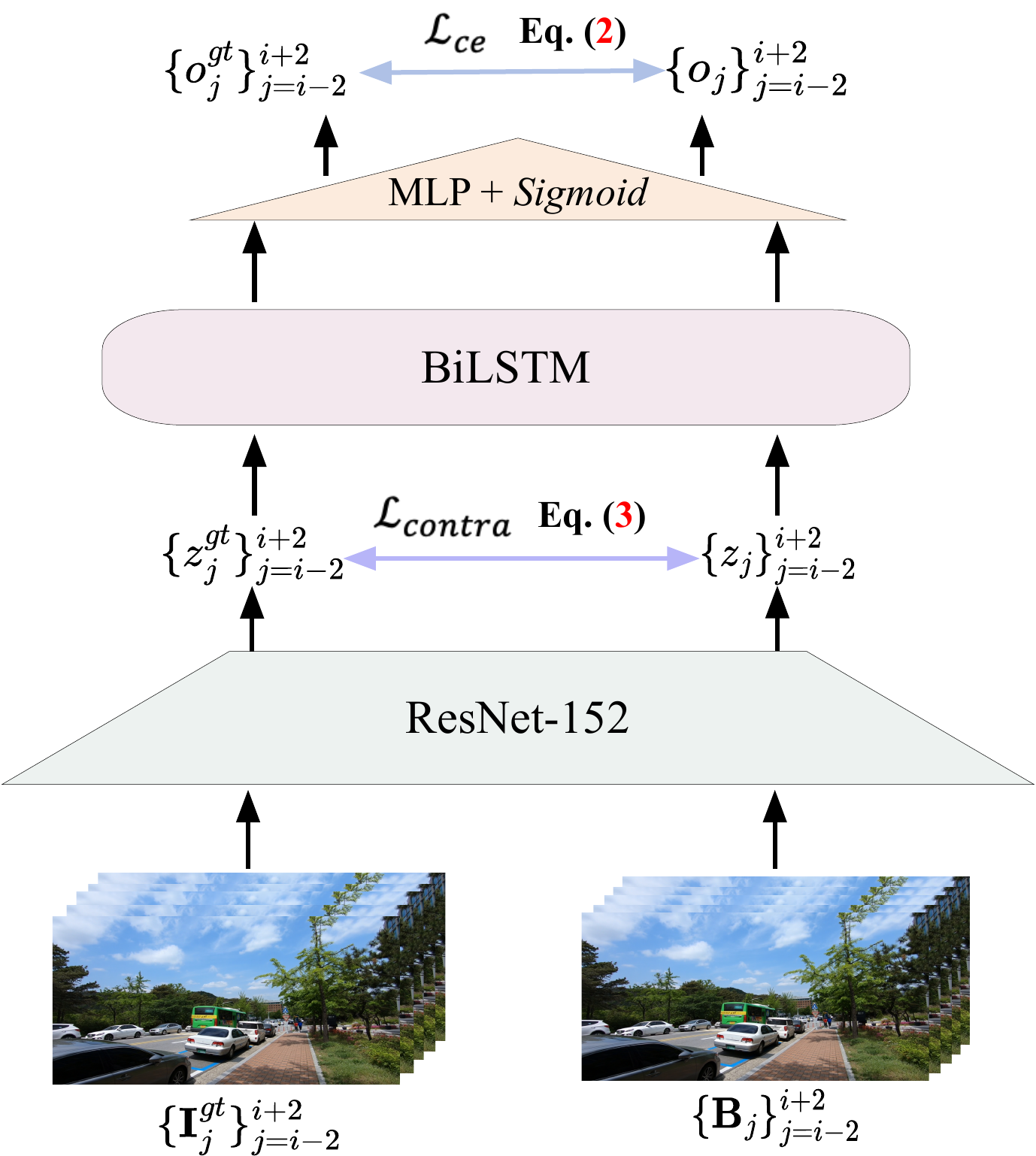} 
		\ \ \
	\end{tabular}
	\caption{The architecture of the blur-aware detector is designed to distinguish between sharp and blurry frames, where a set of 5 adjacent frames, \ie, $\bm{B}_{i-2},\cdots,\bm{B}_{i},\cdots,\bm{B}_{i+2}$, are taken as input. The current frame $\bm{B}_i$ is then classified as either blurry or sharp. 
		During the training process of the detector, we introduce contrastive loss Eq. \eqref{eq: contrast} is introduced to enhance the discrimination between blurry and sharp frames. }
	\label{fig:detector}
\end{figure}

\subsection{Video Deblurring Framework by Aggregating {Nearest} Sharp Features}
For an observed blurry video $\{\bm{B}_i\}_{i=1}^M$ with $M$ frames, video deblurring aims to recover consecutive sharp frames $\{\hat{\bm{I}}_i\}_{i=1}^{M}$.
In existing video deblurring methods~\citep{pan2023cascaded,kim2018spatio,hyun2017online}, a fact that sharp frames may appear in a real-world blurry video is ignored, and thus research attention is mainly paid on the feature fusion of adjacent blurry frames. 
As shown in Fig. \ref{fig:real_example}, sharp frames usually are present near a blurry frame, and thus temporal {nearest} sharp features are available for facilitating the restoration of a blurry frame. 
It is a natural strategy to take these sharp frames as input of reconstruction network.
However, there usually exists considerable temporal motion between current frame $\bm{B}_i$ and sharp frames. %
This phenomenon has also been investigated in~\citep{shang2021bringing}, where optical flow-based warping network is adopted for spatial alignment of sharp and blurry frames, and the features are concatenated simply, possibly suffering from the problem of undesirable artifacts and not fully removing blur (Please see D$^2$Nets in Figs. \ref{fig:gopro} and \ref{fig:reds}).

In this work, we propose a new video deblurring framework for aggregating {nearest} sharp features. 
For a current frame $\bm{B}_i$ with size $ H \times W \times 3$, our proposed video deblurring framework for aggregating {nearest} sharp features can be defined as 
\begin{equation}
	\begin{aligned}
		{\bm{G}_{i}^+, \bm{G}_{i}^-} &= \mathcal{F}_\text{Detector}( \bm{B}_{i-N},\cdots\bm{B}_{i}, \cdots \bm{B}_{i+N}),\\
		% {\bm{G}_{i-1}^+, \bm{G}_{i-1}^-} &= \mathcal{F}_\text{DET}( \bm{B}_{i-1-N},\cdots\bm{B}_{i-1}, \cdots \bm{B}_{i-1+N}),\\
		\hat{\bm{I}}_i &= \mathcal{F}_\text{HybFormer}(\bm{G}_{i}^-,\bm{B}_{i-1},\bm{B}_{i},\bm{B}_{i+1},\bm{G}_{i}^+), 
	\end{aligned}
\end{equation}
where $\mathcal{F}_{\text{Detector}}$ is a blur-aware detector for finding two sharp frames $\bm{G}_i^+$ and $\bm{G}_i^-$ in adjacent $N$ frames, where the former one is in the rear direction and the latter one is in the front direction of blurry frame $\bm{B}_i$. 
When there are more than one sharp frame in the same direction, we choose the nearest sharp frame from $\bm{B}_i$.
When there is no sharp frame in $N$ adjacent frames, $\bm{G}_i^+$ and $\bm{G}_i^-$ are simply set as $\bm{B}_{i+2}$ and $\bm{B}_{i-2}$, respectively. 
That is to say, for a current frame $\bm{B}_i$, we construct a set $\{\bm{G}_{i}^-,\bm{B}_{i-1},\bm{B}_{i},\bm{B}_{i+1},\bm{G}_{i}^+\}$, based on which the blurry frame $\bm{B}_i$ can be restored by hybrid Transformers $\mathcal{F}_\text{HybFormer}$ for aggregating features from neighboring frames $\bm{B}_{i-1}$ and $\bm{B}_{i+1}$, and detected sharp frames $\bm{G}_i^+$ and $\bm{G}_i^-$. 
It is worth noting that in order to maintain generality, we do not have any requirements for whether current and neighboring frames are sharp or blur.
We note that detected sharp frames $\bm{G}_i^+$ and $\bm{G}_i^-$ may still have mild blur, and thus are processed in the same way.  
Therefore, for a blurry video $\{\bm{B}_i\}_{i=1}^M$, deblurring is conducted by processing each item in the set $\{\bm{G}_{i}^-,\bm{B}_{i-1},\bm{B}_{i},\bm{B}_{i+1},\bm{G}_{i}^+\}_{i=1}^{M}$.
The overall algorithm is presented in Alg. \ref{alg:algorithm}. 
Moreover, we propose an event fusion module, based on which the proposed method is easily extended to event-driven video deblurring. 
\begin{algorithm}[t]\small
	\caption{Aggregating Sharp Features for Video Deblurring }
	\label{alg:algorithm}
	\begin{algorithmic}[1]
		\REQUIRE Blurry video with $M$ frames $\{\bm{B}_{i}\}_{i=1}^{M}$ %(and optional events $\{\bm{E}_{i}\}_{i=1}^{M}$)
		\ENSURE  Deblurring video $\{\hat{\bm{I}}_{i}\}_{i=1}^{M}$
		\STATE Distinguishing sharp and blurry frames using $\mathcal{F}_\text{Detector}$. 
		\STATE Constructing a set $\{\bm{G}_{i}^-,\bm{B}_{i-1},\bm{B}_{i},\bm{B}_{i+1},\bm{G}_{i}^+\}_{i=1}^{M}$, where $\bm{G}_{i}^- \text{ and }\bm{G}_{i}^+$ are detected sharp frames of $\bm{B}_i$ in the front and rear directions. 
		\FOR{ $i = 1 : M$ }		
		\STATE $	%\begin{aligned}
		\hat{\bm{I}}_i=\mathcal{F}_\text{HybFormer}(\bm{G}_{i}^-,\bm{B}_{i-1},\bm{B}_{i},\bm{B}_{i+1},\bm{G}_{i}^+).
		$%\end{aligned}
		\ENDFOR
		\RETURN Deblurring video frames $\{\hat{\bm{I}}_{i}\}_{i=1}^{M}$
	\end{algorithmic}
\end{algorithm}

\subsection{Detecting Sharp Frames} \label{sec: detector}

We treat distinguishing sharp and blurry frames in a video as a binary classification task. 
Following~\cite{shang2021bringing}, we consider the temporal information in video and thus adopt bidirectional LSTM (BiLSTM)~\citep{hochreiter2001discrete} to act as the classifier, by which correlations of adjacent frames in both forward and backward directions are leveraged. 
The architecture of {BiLSTM detector} is visualized in Fig. \ref{fig:detector}. 
For a sequence of video frames, {the detector} first extracts features using ResNet-152, and then transforms features to a 512-dimension vector $\{{z}_{i}\}^{M}_{i=1}$ {as the input of BiLSTM},  where $M$ is the total number of frames of the input video.   
Finally, \emph{Sigmoid} function is used to normalize the outputs of BiLSTM in the range [0,1], indicating a frame is blurry or sharp.

For the consecutive frames in a input video $\{\bm{B}_{i}\}^{M}_{i=1}$, the output of detector is denoted by $\{{o}_{i}\}^{M}_{i=1}$, in which ${o}_{i}$ is the probability of $\bm{B}_{i}$ being a sharp frame. 
We binarize the outputs of BiLSTM by threshold $\epsilon = 0.5$. 
A frame $\bm{B}_i$ is blurry, if {$o_i < \epsilon$}, otherwise $\bm{B}_i$ is sharp.  
Then for a given current frame $\bm{B}_i$, we can detect two sharp frames \ie, $\bm{G}_i^-$ and $\bm{G}_i^+$, from its $N$ adjacent frames in the front and rear directions, respectively. 
%
%	In principle, we choose two nearest frames with {$o =1$} as $\bm{G}_i^-$ and $\bm{G}_i^+$ in the front and rear, respectively. 
If sharp frames cannot be found, we simply set $\bm{G}_i^-$ and $\bm{G}_i^+$ as the frames $\bm{B}_{i-2}$ and $\bm{B}_{i+2}$, respectively. 
In this work, we empirically set the searching range {$N = 7$}. 
This is because sharp frames beyond this range may have significant distinctions from the scene content in $\bm{B}_i$, and thus are not suitable to extract {nearest} sharp features. 
{Please refer to Section~\ref{sec:nsf} for the impact of N values.}

To make the training of $\mathcal{F}_\text{Detector}$ easier, we split the video sequence into segments, each of which contains $5$ frames. 
%
%Empirically, we set $n=5$ in this work. 
The blur-aware detector can be trained by minimizing the binary cross-entropy loss function
\begin{equation}
	\mathcal{L}_\text{ce}=-\sum_{j\in \mathbb{N}}\left(o_j^{gt}\log(o_j) + (1-o_j^{gt})\log(1-o_j)\right),
\end{equation}
where $\mathbb{N}$ is the index set of 5 adjacent frames, ${o}_{j}^{gt}$ denotes the label of $j$-th frame, \ie, ${o}_{j}^{{gt}} = 1$ when $\bm{B}_{j}$ is a sharp frame, otherwise ${o}_{j}^{{gt}} = 0$. 
In this way, temporal information in 5 adjacent frames is considered when distinguishing sharp and blurry frames. 
However, we find that the trained detector is not well generalized to real-world blurry videos, by only adopting the binary cross-entropy loss function. 
And thus, we further introduce a supervised contrastive loss function
\begin{equation}\label{eq: contrast}
	\mathcal{L}_\text{contra}\!=\!-\log\!\left(\frac{\sum\limits_{j\in \mathbb{N}^s} \exp(\bm{z}_{j} \cdot \bm{z}_j^{gt})}{\sum\limits_{j\in \mathbb{N}^s}\!\!\!\exp(\bm{z}_{j} \cdot \bm{z}_j^{gt}) + \!\!\!\!\sum\limits_{j \in \mathbb{N}^b} \!\!\! \exp(\bm{z}_{j} \cdot \bm{z}_j^{gt})}\right),
\end{equation} 
where $\mathbb{N}^s$ and $\mathbb{N}^b$ are subsets of $\mathbb{N}$, satisfying that $j\in \mathbb{N}^s$ if ${o}_{j}^{{gt}} = 1$, otherwise $j\in \mathbb{N}^b$. 
Then $\bm{z}_{j}^{gt}$ denotes the feature vector of ground-truth sharp image and can be obtained by passing through the same ResNet-152 for feature extraction, and the symbol '$\cdot$' denotes the normalized inner dot product.
Benefiting from contrastive loss $\mathcal{L}_\text{contra}$, the features between sharp and blurry frames are forced to be as distinguishable as possible, leading to better generalization ability on real-world blurry videos. 

Finally, the total loss function for training $\mathcal{F}_\text{Detector}$ is defined as 
\begin{equation}
	\mathcal{L}_\text{Dector}= \mathcal{L}_\text{ce} + \lambda \mathcal{L}_\text{contra},
\end{equation}
where $\lambda$ is a hyper-parameter and is set $\lambda = 10$ in all the experiments.

\subsection{Video Deblurring via Hybrid Transformers}\label{sec:hybformers}
\begin{figure*}[!t]\footnotesize
	%	\centering
	%\fbox{\rule{0pt}{3in} \rule{0.9\linewidth}{0pt}}
	\begin{tabular}{lcccccc}
		\ \ \
		\includegraphics[width=\linewidth]{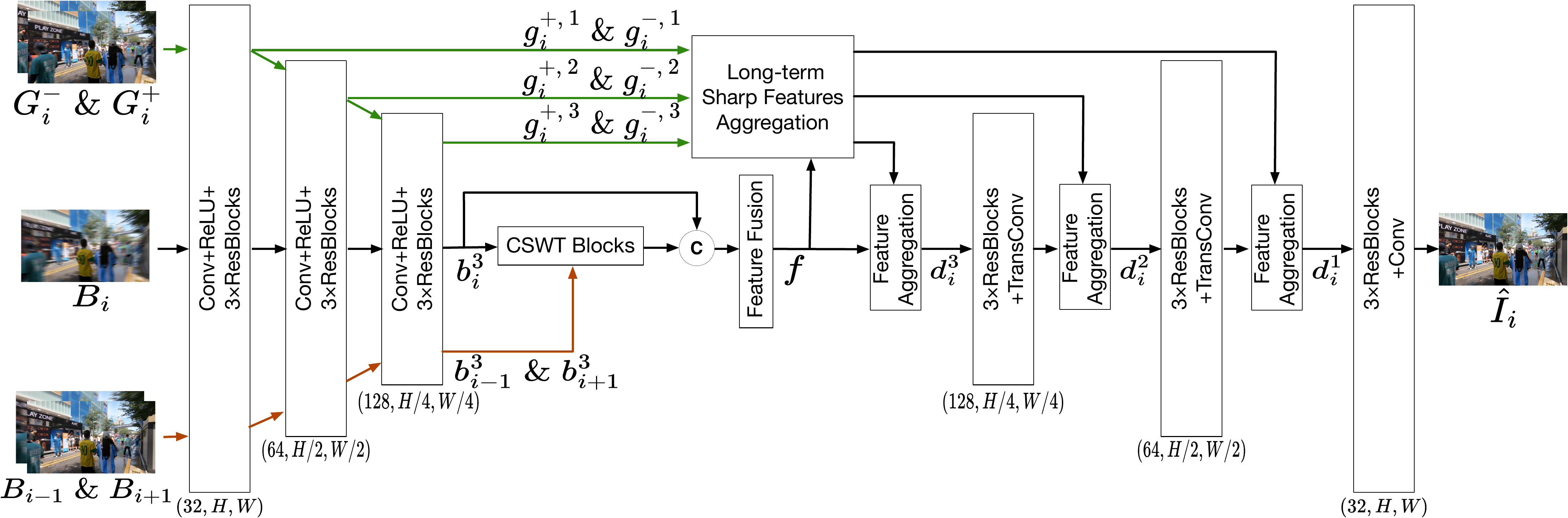} 
		\ \ \
	\end{tabular}
	\caption{{ The flowchart of $\mathcal{F}_\text{HybFormer}$ for restoring a blurry frame $\bm{B}_i$ with its neighboring frames $\bm{B}_{i-1}$ and $\bm{B}_{i+1}$, as well as its corresponding detected sharp frames $\bm{G}_{i}^- \text{ and } \bm{G}_{i}^+$, is presented. 
			The restoration process in $\mathcal{F}_\text{HybFormer}$ involves four steps: extracting features from all frames using a three-scale CNN, fusing the features of adjacent frames by cross-attention shifted window Transformer (CSWT) blocks,  aggregating {nearest} sharp features using global Transformer, and finally reconstructing the latent frame $\hat{\bm{I}}_i$ with a decoder based on a three-scale CNN. 
			Details regarding the window-based local Transformer and global Transformer can be found in Figs. \ref {fig: cswt} and \ref {fig: msm}, respectively.
	} }
	\label{fig:restor}
\end{figure*}
The hybrid Transformer $\mathcal{F}_\text{HybFormer}$ mainly consists of two key components, \ie, window-based local Transformer for aggregating features from neighboring frames $\bm{B}_{i-1}$ and $\bm{B}_{i+1}$, and global Transformer for aggregating {nearest} sharp features from detected sharp frames $\bm{G}_i^+$ and $\bm{G}_i^-$. 
As shown in Fig. \ref{fig:restor}, hybrid Transformers are implemented in the feature space extracted by CNN. 
In particular, a three-scale CNN encoder is adopted for feature extraction, where an input frame with size $ H \times W\times 3$ is transferred to features with three scales $H \times W \times C$, $ \frac{H}{2} \times \frac{W}{2} \times 2C$, and  $\frac{H}{4} \times \frac{W}{4} \times 4C$, where $C$ is channel number and we set $C=32$. 
The window-based local Transformer is implemented in the third scale, while global Transformer is implemented in multi-scale scheme. 
Finally, the aggregated features are reconstructed to sharp frames in a corresponding three-scale CNN decoder. 

As for learning the parameters of $\mathcal{F}_\text{HybFormer}$, we adopt $\ell_1$-norm loss function
\begin{equation} \label{eq:enh}
	\begin{aligned}
		\hat{\bm{I}}_i = \mathcal{F}_\text{HybFormer}&(\bm{G}_{i}^-,\bm{B}_{i-1},\bm{B}_{i},\bm{B}_{i+1},\bm{G}_{i}^+), \\
		\mathcal{L}_\text{HybFormer}&=\left\|\hat{\bm{I}}_i-\bm{I}_{i}^{gt}\right\|_{1}.
	\end{aligned}
\end{equation}

\subsubsection{Window-based Local Transformer for Aggregating Neighboring Frames}\label{sec:cswt} 
\begin{figure*}[!t]\footnotesize
	\centering
	%\fbox{\rule{0pt}{3in} \rule{0.9\linewidth}{0pt}}
	\begin{tabular}{lcccccc}
		\ \ \
		\includegraphics[width=0.82\linewidth]{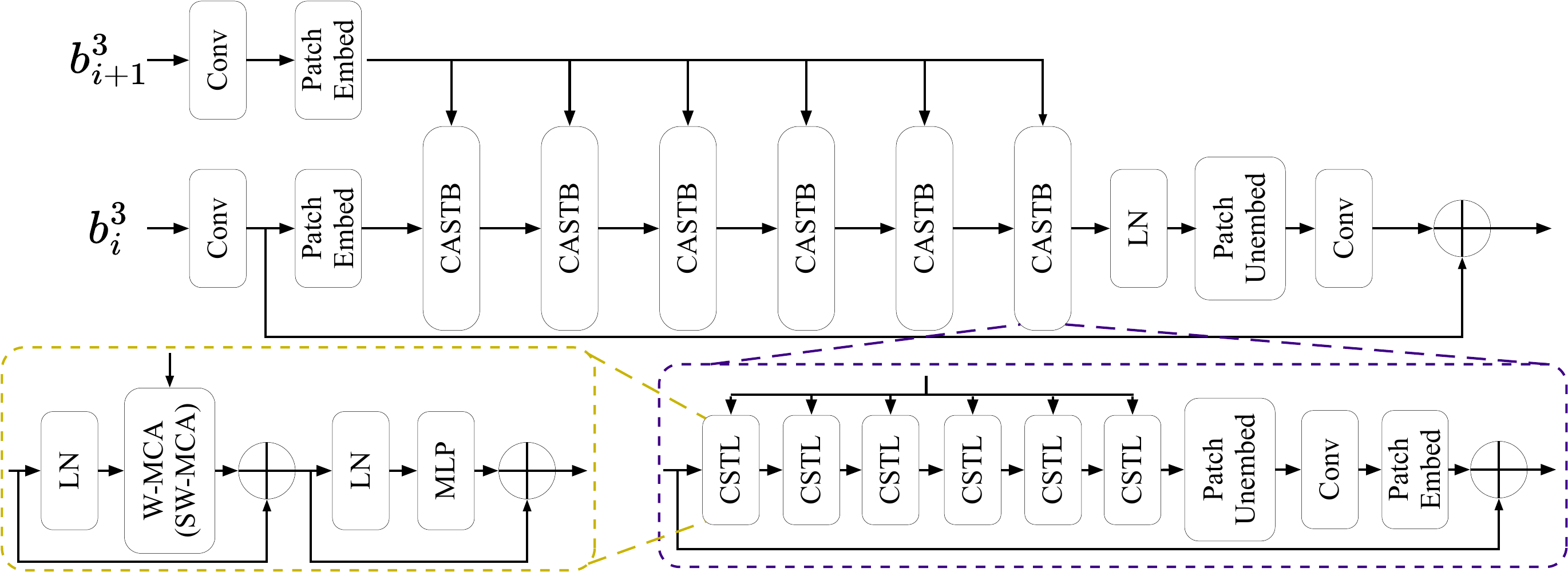} 
		\ \ \
	\end{tabular}
	\caption{The architecture of CSWT Blocks $f_\text{cswt}$ for fusing adjacent frame features, where the third scale features $\bm{b}^3_i$ and $\bm{b}^3_{i+1}$ of frames $\bm{B}_i$ and $\bm{B}_{i+1}$ can be aggregated using cross-attention without explicit spatial alignment. 
	} 
	\label{fig: cswt}
\end{figure*}

As shown in Fig. \ref{fig: cswt}, we adopt cross-attention shifted window Transformer (CSWT) blocks $\bm{f}_\text{cswt}$ to fuse current frame $\bm{B}_i$ and its two neighboring frames $\bm{B}_{i-1}$ and $\bm{B}_{i+1}$, which is implemented in the third scale. 

We take the feature aggregation of $\bm{B}_i$ and $\bm{B}_{i+1}$ as example, and $\bm{B}_{i-1}$ can be processed in the same way. 
Let $\bm{b}_i^3$ and $\bm{b}_{i+1}^3$ be the features of $\bm{B}_i$ and $\bm{B}_{i+1}$ in third scale with size ${ \frac{H}{4} \times \frac{W}{4} \times 16C}$. 
The key idea of window-based local Transformer $f_\text{cswt}$ is cross-attention without spatial alignment, where the features extracted from $\bm{b}_{i}^{3}$ as $Key$ and $Value$ matrices, and features extracted from $\bm{b}_{i+1}^{3}$ as $Query$ matrices. 
The architecture of $f_\text{cswt}$ is borrowed from~\citep{liang2021swinir,liu2021Swin}, and self-attention is modified to cross-attention for integrating adjacent frames. 
The architecture of $\bm{f}_\text{cswt}$ can be seen in Fig. \ref{fig: cswt}. 
Following the setting in SwinIR~\citep{liang2021swinir}, there are 6 Cross-Attention Swin Transformer Blocks (CASTB) in $\bm{f}_\text{cswt}$. 
Each CASTB contains 6 Cross Swin Transformer Layers (CSTL) and 1 convolutional layer. 
For each CSTL, it includes a LayerNorm(LN), W-MCA or SW-MCA and MLP, where W-MCA and SW-MCA denote window based multi-head cross-attention using regular and shifted window partitioning configurations, respectively. 
W-MCA and SW-MCA alternate in each CSTL, which maintain the efficient computation of non-overlapping windows and improve modeling power, as described in~\citep{liu2021Swin}. 
The number of attention heads is 8. 

Then, the features from two neighboring frames can be aggregated as 
\begin{equation} \label{eq:swt}
	\begin{aligned}
		\bm{b}'_{i-1}&=f_{\text{cswt}}\left(\bm{b}_{i}^{3}, \bm{b}_{i-1}^{3}\right), \\
		\bm{b}'_{i+1}&=f_{\text{cswt}}\left(\bm{b}_{i}^{3}, \bm{b}_{i+1}^{3}\right), \\
		\bm{f}&={conv}(\mathcal{C}(\bm{b}'_{i-1}, \bm{b}_{i}^{3},\bm{b}'_{i+1})),
	\end{aligned}
\end{equation}
where $\mathcal{C}$ is concatenation, and $conv$ is a convolutional layer.  
We fed the concatenation of $\bm{b}_{i}^{3}, \bm{b}'_{i-1} \text{ and } \bm{b}'_{i+1}$ into a convolutional layer to get new features $\bm{f}$ which integrates current and neighboring frame for reconstructing the sharp frame in decoder.

% \subsubsection{Matching \& Selection Module} \label{sec: msm}
\subsubsection{Global Transformer for Aggregating Sharp Frames} \label{sec: msm}
\begin{figure*}[!t]\footnotesize
	\centering
	%\fbox{\rule{0pt}{3in} \rule{0.9\linewidth}{0pt}}
	\begin{tabular}{lcccccc}
		\ \ \
		\includegraphics[width=0.9\linewidth]{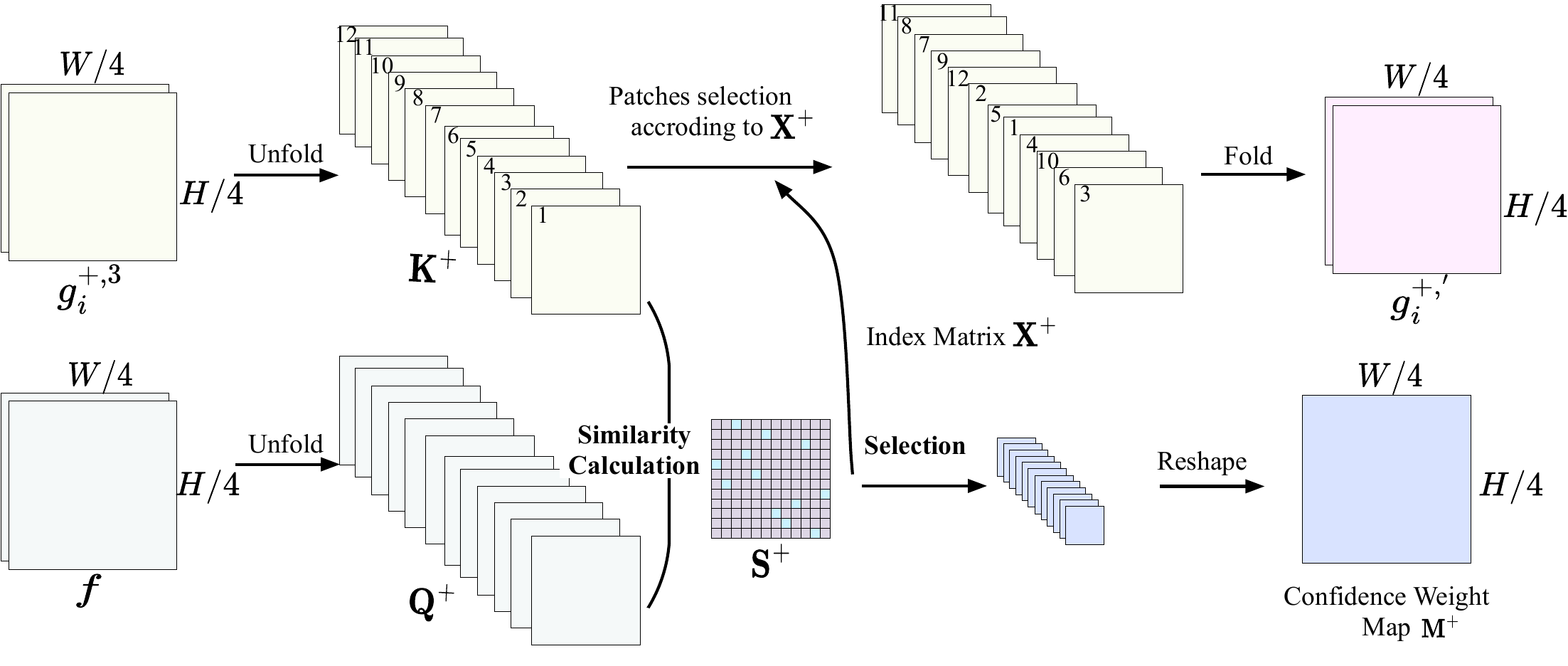} 
		\ \ \
	\end{tabular}
	\vspace{-1em}
	\caption{Global attention of {nearest} sharp feature $\bm{g}^{+,3}_i$ in rear frame $\bm{G}^+_i$ and feature $\bm{f}$ in the third scale. 
		The global similarity for each patch is recorded in $\bm{S}^+$, enabling identification of the most relevant patches in the index matrix $\bm{X}^+$ along with their corresponding confidence map $\bm{M}^+$. These patches are then folded into a new feature representation denoted as $\bm{g}_i^{+,'}$. Similarly, the {nearest} sharp features $\bm{g}_i^{-,3}$ from the front frame $\bm{G}_i^-$ can be processed in the same way. 
	} 
	\label{fig: msm}
\end{figure*}

Considering the temporal motion between current frame and detected sharp frames, we propose to utilize global attention across current frame and sharp frames to measure the similarity for each coordinate. 
The key idea of global Transformer includes a matching operation for calculating relevance between current frame and sharp frames, and a selection operation for picking out the most relevant {nearest} sharp information to the current frame.  
We take the aggregation of {nearest} features from rear sharp frame $\bm{G}_i^+$ as an example to show how to compute global attention. 

\textbf{\emph{Global Attention for Aggregating {Nearest} Sharp Features.}}
We need to compute the global similarity between feature $\bm{f}$ and feature $\bm{g}_i^+$ of detected sharp frame $\bm{G}_i^+$. 
By taking $\bm{g}_i^{+,3}$ in third scale as an example, we show how to compute the global similarity. 
The features $\bm{f}$ and $\bm{g}_i^{+,3}$ are unfolded into patches to form $Query$ matrix $\bm{Q}^+$ with size $ \frac{HW}{16} \times p^2$ and $Key$ matrix $\bm{K}^+$ with size $p^2 \times \frac{HW}{16}$, and we normalize $\bm{Q}^+$ and $\bm{K}^+$ to obtain the similarity matrix $\bm{S}^+ = \bm{Q}^+\times \bm{K}^+$ with size $\frac{HW}{16} \times \frac{HW}{16}$, whose element measures the similarity between extracted patches. 

For a patch $i$, the most relevant patch in sharp feature can be found by $\bm{X}^+(i) = \arg \underset{}{\max}\, \bm{S}^+(i,:)$ as well as confidence weight map $\bm{M}^+(i)=  \underset{}{\max}\, \bm{S}^+(i,:)$,
where $\bm{M}^+$ and $\bm X^+$ with size $\frac{HW}{16} \times 1$ are the index to the most similar patches and the confidence weight map. 
Then, $\bm{M}^+$ and $\bm X^+$ can be reshaped to matrices with size $\frac{H}{4} \times \frac{W}{4}$, having the same dimension with features in third scale. 

Next, we can get the sharp features that are most related to the current frame patch according to index matrix $\bm X^+$ and fold these patches together, which is the reverse operation of the unfolding operation to get new features $\bm{g}_{{i}}^{+,'}$. 
Instead of directly applying the confidence weight map $\bm{M}^+$ to $\bm{g}_i^{+,3}$,
we first fuse the sharp texture features $\bm{g}_i^{+,3}$ from detected sharp frame with features $\bm{f}$ to leverage more information. Such fused features are further element-wisely multiplied by the confidence weight map $\bm M^+$ and added back to $\bm{f}$ to get the final input of $f_\text{dec}$.
This step can be represented as 
% \begin{equation}
	\begin{gather}
		\bm{g}_{i}^{+,'} = f_{\text{fold}}(\bm{g}_{{i}}^{+,3}; \bm{X}^+ ),\nonumber\\
		\bm{g}_{i}^{-,'} = f_{\text{fold}}(\bm{g}_{{i}}^{-,3}; \bm{X}^- ),\nonumber\\
		{\bm{d}}_i^3\!=\!conv\left(\mathcal{C}(\bm{g}_{i}^{+,'}, \bm{f})\right) \!\odot\! \bm{M}^+ \nonumber\\
		\!\!\!+ conv\left(\mathcal{C}(\bm{g}_{i}^{-,'}, \bm{f})\right)\! \odot \!\bm{M}^-
		\!\!\!+ \!\bm{f},
	\end{gather}
	% \end{equation}
where $f_{\text{fold}}$ denotes searching most relevant patches according to index $\bm{X}^+$ to get folded new features $\bm{g}_{{i}}^{+,'}$, and $\odot$ denotes element-wise multiplication.
% up/down-sampling
And the features $\bm{g}_i^{-,'}$ from front sharp frame $\bm{G}_i^-$ can be handled in the same way. 
Finally, the {nearest} sharp features can be aggregated with feature $\bm f$, where confidence maps $\bm M^+$ and $\bm M^-$ are considered when aggregating features.

\textbf{\emph{Implementation in Multi-scale Scheme.}}
To fully exploit {nearest} sharp features, feature aggregation is performed in the multi-scale scheme, where features in first and second scales $\bm{g}^{+,2}_i$ and $\bm{g}^{+,1}_i$ are also aggregated. 
Due to the computational cost of global attention, we do not suggest to directly compute index matrix $\bm X^+$ and confidence map $\bm M^+$ in the first and second scales. 
As for the confidence map $\bm M^+$, it can be upsampled with corresponding factor. 
And for index matrix $\bm X^+$, we need to keep the same dimension for unfolded feature tensors in first and second scales. 
Therefore, to ensure the applicability of $\bm{X}^+$ at different scales, we adjusted the parameters of unfolding operation at different scales to ensure that the total numbers of unfolding patches in three scales are consistent.

%
% In this section, we will elaborate on the unfolding operation on features with different scales.
Recalling that the spatial size of $\bm X^+$ is $\frac{H}{4}\times \frac{W}{4}$, \ie, the total number of extracted patches is $L = \frac{H}{4}\times \frac{W}{4}$. 
Therefore, we need to keep the same number for the other two scales. 
We have got the {nearest} features in three-scale CNN, \ie, $\bm{g}_{{i}}^{+,1}$, $\bm{g}_{{i}}^{+,2}$ and $\bm{g}_{{i}}^{+,3}$, whose spatial sizes are $H \times W$, $\frac{H}{2} \times \frac{W}{2}$ and $\frac{H}{4} \times \frac{W}{4}$. 
Unfolding operation aims to extract sliding local patches from a batched input tensor. 
The total number of sliding window patches is
\begin{equation} \label{eq: unfold}
	L=\prod_{x\in\{h,w\}}\left\lceil\frac{x +2 \times p - (k-1)}{r}\right\rceil
\end{equation}
where $\lceil \cdot \rceil$ is the ceiling operation, $h$ and $w$ are height and width of features, and ${p}$, ${k}$, ${r}$ denote the padding size, patch size and stride size, respectively. 
To satisfy the condition $L = \frac{H}{4}\times \frac{W}{4}$, we set the parameters 
$\{[h,w], [k,k], [p,p],[r,r]\}$ for the features $\bm{g}_i^{+,3}$, $\bm{g}_i^{+,2}$, $\bm{g}_i^{+,1}$ in three scales respectively as 
$\{[\frac{H}{4},\!\frac{W}{4}],\![3,\!3],\![1,\!1],\![1,\!1]\}$,
$\{[\frac{H}{2},\!\frac{W}{2}],\![6,\!6],\![2,\!2],\![2,\!2]\}$,
$\{[{H},\!{W}],\![12,\!12],\![4,\!4],\![4,\!4]\}$. 
And the features $\bm{g}^-_i$ share the same setting. 
In addition, the center spatial position of the same patch with different scales keeps unchanged, guaranteeing that the confidence weight map $\bm M$ can be upsampled for different scales by bilinear interpolation. 

Then, the feature aggregation in the second scale can be defined as 
% \begin{equation}
	\begin{gather}
		\bm{d}^{'}= f_\text{dec}({\bm{d}_i^3}),\nonumber \\  %\uparrow 
		\bm{g}_{i}^{+,'} = f_{\text{fold}}(\bm{g}_{{i}}^{+,2}; \bm{X}^+ ),\nonumber\\
		\bm{g}_{i}^{-,'} = f_{\text{fold}}(\bm{g}_{{i}}^{-,2}; \bm{X}^- ),\nonumber\\
		\bm{d}_i^2\!=\!conv\!\left(\mathcal{C}(\bm{g}_{i}^{+,'}, \bm{d}^{'})\right) \!\odot\! \bm{M}^+ \!\!\uparrow\!\! \nonumber\\ 
		+ conv\!\left(\mathcal{C}(\bm{g}_{i}^{-,'}, \bm{d}^{'})\right)\! \odot \!\bm{M}^- \!\!\uparrow\!\!+ \!\bm{d}^{'},
	\end{gather}
	where $f_\text{dec}$ is one layer in decoder, consisting of three ResBlocks and one transposed convolution, and the feature and confidence map are upsampled with factor 2. 
	The aggregation in the scale 1 shares similar steps, where the feature $\bm{d}_i^1$ is with size $H\times W \times C$ and is finally mapped to the reconstructed sharp frame $\hat{\bm{I}}_i$ with size $H\times W \times 3$ using three ResBlocks and one convolutional layer. 
	
	\subsection{Extension to Event-driven Video Deblurring } \label{sec:event}
	We take one step further to leverage events in $\mathcal{F}_\text{HybFormer}$ for event-driven deblurring. 
	For an event camera, given a current frame $\bm{B}_i$, its corresponding stream of events $\bm{E}_i$ are available.  
	Each event has the form {$\mathbf(t,x,y,p)$}, which records intensity changes for coordinates $\mathbf(x,y)$ at time ${t}$, and polarity $p = \pm 1$ denotes the increase or decrease of intensity change.  
	In this work, we transform the events stream into a tensor with 40 channels for each frame as in~\citep{wang2020event}. 
	We introduce an event fusion module to incorporate events into frame features for event-driven deblurring. %
	Our event fusion module transfers the event $\bm{E}_i$ to feature $\bm e$ through 5 convolution layers, having the same size as $\bm f$. 
	Event feature $\bm e$ can be upsampled for different scales by bilinear interpolation. 
	Then an fusion module is implemented by calculating a reweighting map through cross-attention between event feature and frame feature, which can be used to facilitate deblurring by matrix multiplication with the features of frames.  
		\begin{table*}[!tb]\small
		\setlength{\tabcolsep}{4pt}
		\caption{Quantitative comparison of deblurring results of on GoPro dataset. 
		}
		\centering
		\resizebox{\columnwidth}{!}{
			\begin{tabular}{c|ccccccc|c}
				\hline
				
				\hline
				Method &  DMPHN &  STFAN & 
				% CDVD-TSP\citep{pan2020cascaded}  & 
				ESTRNN  &  MIMO-UNet++  & D$^2$Nets & CDVD-TSPNL & VRT   & Ours  \\
				\hline
				PSNR & 32.09   & 31.76  &    
				% 33.76     & 
				33.52  &  35.49  &35.18  & 36.65   & {37.02}  &  \textbf{37.33}      \\
				SSIM & 0.897  & 0.873  &   
				% 0.925       &
				0.912  &   0.938  &0.943  & 0.958 & {0.961}  & \textbf{0.962}         \\
				
				\hline
				
				\hline
			\end{tabular}
		}
		\label{tabel:gopro-vi}
		\vspace{-0.1in}
	\end{table*}
	\begin{figure*}[!htb]\footnotesize
		\centering
		\setlength{\tabcolsep}{0pt}
		
		\begin{tabular}{cclcclcclcclcclccl}
			\includegraphics[width=0.99\linewidth]{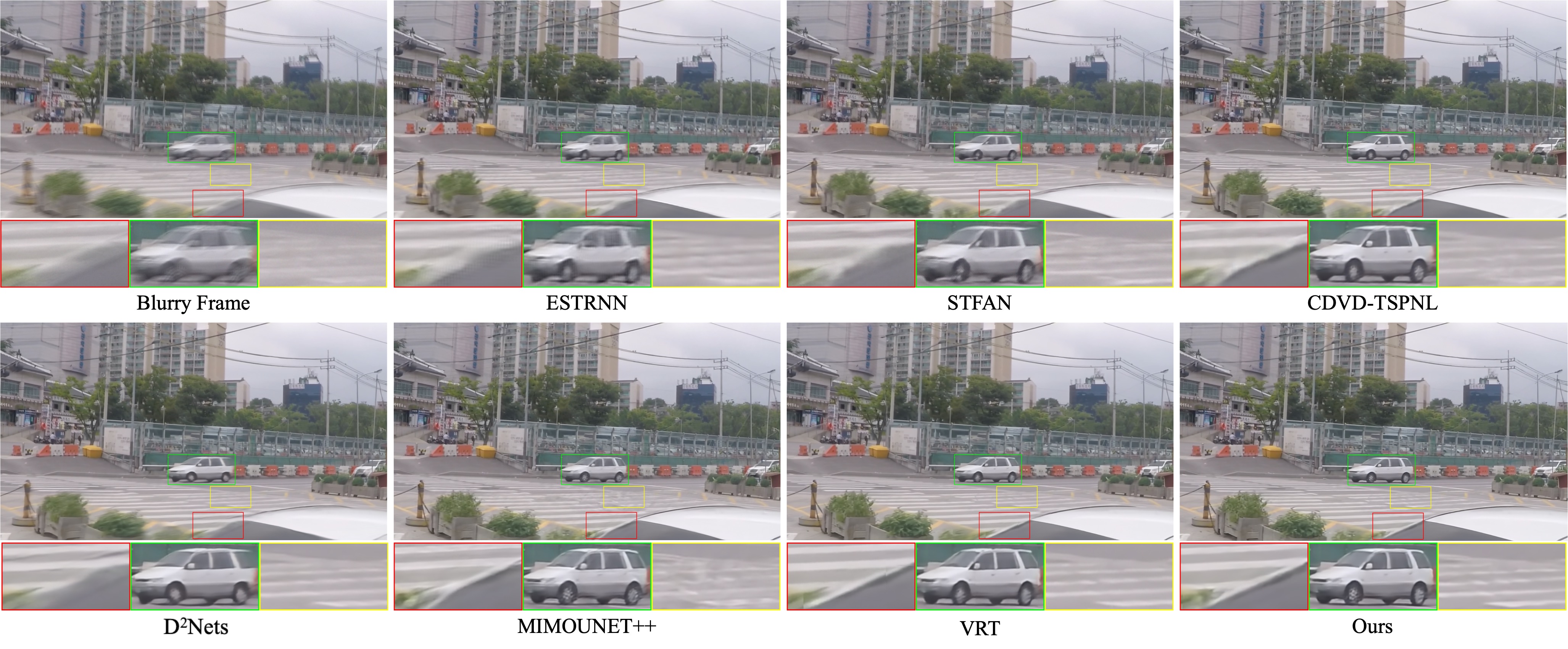}\\
			
		\end{tabular}
		\vspace{-1em}
		\caption{ Visual comparison of deblurring results on GOPRO dataset. }
		\label{fig:gopro}
	\end{figure*}
	\section{Experiments}\label{sec:experiments}
	In this section, we evaluate our proposed method on five benchmark datasets, including GOPRO~\citep{nah2017deep}, REDS~\citep{nah2017deep} and BSD~\citep{zhong2023real} datasets for video deblurring, and RBE~\citep{pan2019bringing} and CED~\citep{scheerlinck2019ced} datasets for event-driven video deblurring. And we conduct ablation study to verify the key contributions of our proposed method on GOPRO dataset.
	As for state-of-the-art competing deblurring methods, we take both image deblurring and video deblurring methods into comparison, where image deblurring methods include DMPHN~\citep{zhang2019deep} and MIMO-UNet++~\citep{cho2021rethinking}, and video deblurring methods include STFAN~\citep{zhou2019spatio}, 
	% CDVD-TSP~\citep{pan2020cascaded},
	ESTRNN~\citep{zhong2023real}, CDVD-TSPNL~\citep{pan2023cascaded}, VRT~\citep{liang2024vrt} and D$^2$Nets~\citep{shang2021bringing}. 
	As for event-driven deblurring methods, we take BHA~\citep{pan2019bringing}, eSL-Net~\citep{wang2020event}, eSL-Net++\citep{yu2022learning} and D$^2$Nets*~\citep{shang2021bringing} into comparison. 
	Finally, we validate their generalization performance on real-world blurry videos. More video results can be found from the link\footnote{\url{https://1drv.ms/f/s!AtY7eoZyJkNBhESrpnW3-IyL9-iy?e=VgvlN1}}.

	\subsection{Datasets and Training Details}

	\subsubsection{Datasets}\label{sec: data}
	For video deblurring, three datasets are adopted for evaluation.
	{The GOPRO and REDS datasets, which comprise synthetically generated blurry videos. In contrast, the BSD dataset, which contains recordings of real-world scenarios, provides a more challenging evaluation ground that reflects the variability and complexity of actual blurry frames.} For event-driven video deblurring, synthetic blurry dataset CED and real-world blurry dataset RBE are adopted.
	
	\textbf{GOPRO Dataset:} GOPRO dataset~\citep{nah2017deep} is widely adopted for image and video deblurring. 
	The videos in the original GOPRO dataset are with framerate of 240 FPS. We follow~\citep{nah2017deep} to split the training and testing sets.
	To satisfy the condition that sharp frames exist in a blurry video, we generate non-consecutively blurry frames in a video by randomly averaging adjacent sharp frames, \ie, the average number is randomly chosen from 1 to 15. 
	And we classify that a generated frame $\bm{B}_i$ is sharp if the number of averaging frames is smaller than 5, \ie, $o_i^{gt}=1$, otherwise $o_i^{gt}=0$. 
	It is worth noting that we randomly generate 50$\%$ blurry frames in a video, while the other 50$\%$ frames are sharp, without constraining that there must be 2 sharp ones in consecutive 7 frames.

	\textbf{REDS Dataset:} 
	Another widely used dataset is REDS dataset~\citep{nah2017deep}, and there are 240 videos for training, 30 videos for testing.
	We generate non-consecutively blurry frames in a similar way as GOPRO. 
	{But the framerate of REDS} is not high enough, simply averaging frames may generate unnatural spikes or steps in the blur trajectory~\citep{nah2017deep}, especially when the resolution is high and the motion is fast. 
	Hence, we employed FLAVR~\citep{kalluri2020flavr} to interpolate frames, increasing the framerate to virtual 960 FPS by recursively interpolating the frames. 
	Thus, we can synthesize frames with more severe degrees of blur, \ie, the average number is randomly chosen from 3 to 39. 
	And we classify that a generated frame $\bm{B}_i$ is sharp if the number of averaging frames is smaller than 17, \ie, $o_i^{gt}=1$, otherwise $o_i^{gt}=0$.
	
	\textbf{BSD Dataset:}
	\citep{zhong2023real} provided a real-world blurry video dataset by using a beam splitter system~\citep{jiang2019learning} with two synchronized cameras. By controlling the length of exposure time and strength of exposure intensity when capturing videos, the system could obtain a pair of sharp and blurry video samples by shooting videos at the same time. 
	They collected blurry and sharp video sequences for three different blur intensity settings, \ie, sharp exposure time - blurry exposure time are set as 1ms-8ms, 2ms-16ms and 3ms-24ms, respectively.
	We note that blurry videos in BSD dataset, especially for long exposure time, suffer from severe blur, but sharp frames still are present to provide {nearest} sharp features. 
	The testing set has 20 video sequences with 150 frames in each intensity setting. We use these testing sets for evaluating generalization ability. 

	\textbf{CED Dataset:}
	\citep{scheerlinck2019ced} presented the first Color Event Camera Dataset (CED) by color event camera ColorDAVIS346, containing 50 minutes of footage with both color frames and events. We also employed FLAVR~\citep{kalluri2020flavr} to interpolate frames for generating blurry frames as the same with REDS. We randomly split the sequences in CED into training, validation and testing sets, and report the corresponding comparison results against the state-of-the-art models by retraining them with the same setting.

	\textbf{RBE Dataset:}
	\cite{pan2019bringing} presented a real blurry event dataset, where each real sequence is captured with the DAVIS under different conditions, such as indoor, outdoor scenery, low lighting conditions, and different motion patterns (e.g., camera shake, objects motion) that naturally introduce motion blur into the APS intensity images. There is no ground-truth data available on this dataset. Hence, we only use it for qualitative comparison.

	\begin{table*}[!tb]\small
		\setlength{\tabcolsep}{4pt}
		\caption{Quantitative comparison of deblurring results on REDS dataset. 
		}
		\centering
		\resizebox{\columnwidth}{!}{
		\begin{tabular}{c|ccccccc|c}
			\hline
			
			\hline
			Method &  DMPHN &  STFAN & 
			% CDVD-TSP\citep{pan2020cascaded}  & 
			ESTRNN  &  MIMO-UNet++  & D$^2$Nets & CDVD-TSPNL &  VRT    & Ours  \\
			\hline
			PSNR &  37.62  & 38.09  &
			% 36.88     & 
			38.38    &   39.05  &  38.82 & 40.06 & {40.84} &  \textbf{41.45}  \\
			SSIM &   0.956  &  0.959 &    
			% 0.955    &
			0.962    &  0.965  & 0.969  & 0.977 &  {0.981}  &  \textbf{0.982}  \\
			
			\hline
			
			\hline
		\end{tabular}
	}
		\label{tabel:reds-vi}
		\vspace{-0.1in}
	\end{table*}
	
	\subsubsection{Training Details}
	In the training process, we use ADAM optimizer~\citep{kingma2015adam} with parameters $\beta_1 = 0.9$, $\beta_2 = 0.999$, and $\epsilon = 1\times 10^{- 8}$ for all the networks in $\mathcal{F}_\text{Detector}$ and $\mathcal{F}_\text{HybFormer}$.
	The batch size is 12 and patch size is set as 200 $\times$ 200. 
	%
	% In order to save training time, we use parameters trained on GOPRO to initialize when training on REDS.  
	%
	The learning rate for reconstruction network $\mathcal{F}_\text{HybFormer}$ is initialized to be $1\times 10^{- 4}$ and is decreased by multiplying 0.5 after every 200 epochs. 
	The training ends after 500 epochs. 
	For BiLSTM detector $\mathcal{F}_\text{Detector}$, the learning rate is set to be $1\times 10^{- 4}$, and the training ends after 20 epochs.  
	%	At blurry restoration stage, we use 3 frames to generate one deblurred frame. In the other stages, we use 5 frames.
	%
	%	Our method is implemented based on PyTorch. 

	\begin{figure*}[!htb]\footnotesize
		\centering
		\setlength{\tabcolsep}{0pt}
		
		\begin{tabular}{cclcclcclcclcclccl}
			\includegraphics[width=0.99\linewidth]{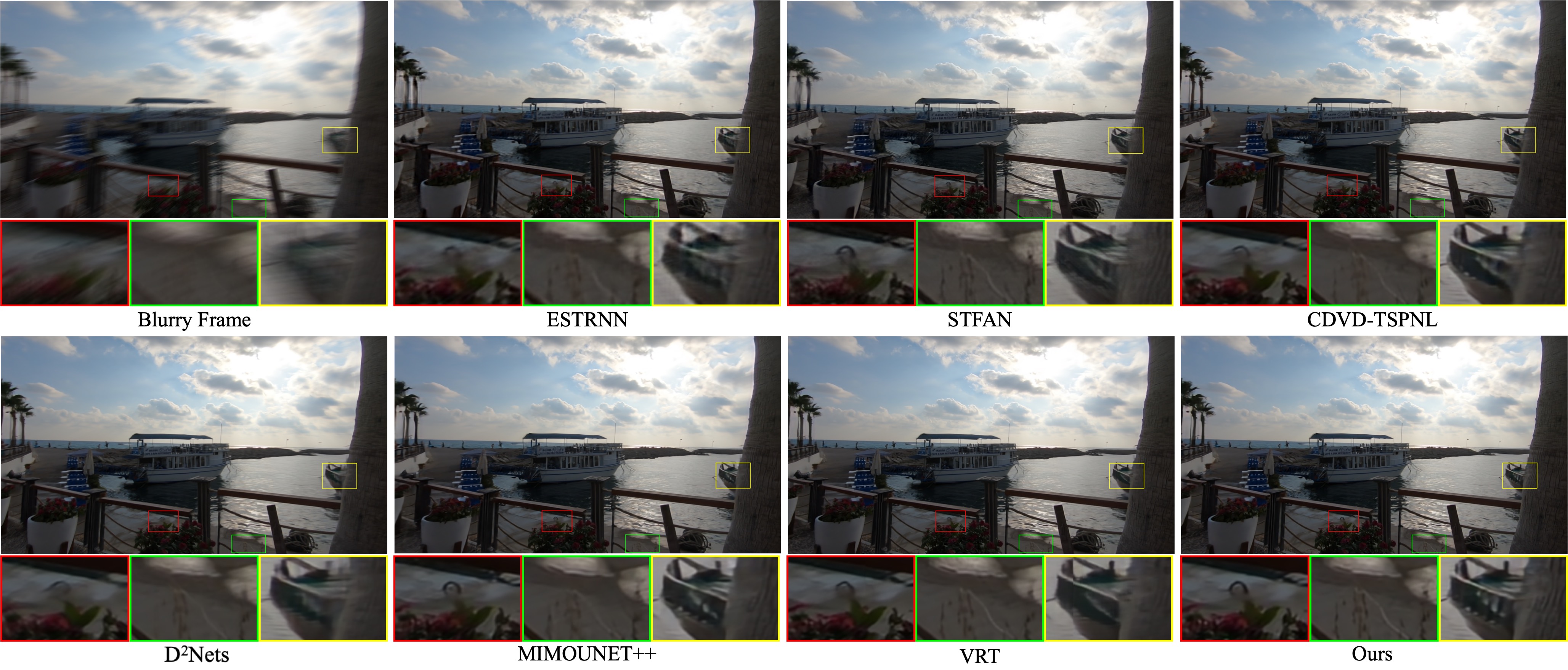}\\

		\end{tabular}
		\vspace{-0.8em}
		\caption{Qualitative comparison on REDS dataset. }
		\label{fig:reds}
	\end{figure*}
	
	\begin{table}[!tb]\small
		\setlength{\tabcolsep}{2.5pt}
		\caption{Comparison of efficiency and model size. The experiment was conducted on GeForce RTX 3090 with videos (240 $\times$ 240). }
		\centering
		\resizebox{\columnwidth}{!}{
		\begin{tabular}{c|ccccccc|c}
			\hline
			
			\hline
			Method &  DMPHN &  STFAN &  
			ESTRNN  &  MIMO-UNet++  & D$^2$Nets & CDVD-TSPNL &  VRT    & Ours \\
			\hline
			Params(M) & 21.70 &  5.37 &  
			2.47  &  16.11   & 32.54  &  5.50  &  35.60    & 30.63\\
			GFLOPs  & 189.3  & 202.9 &  
			11.2  &  135.6 & 326.4 & 1343.9 &  3229.7    & 342.6\\ 
			Runtime(s) &  0.058  & 0.150 &  
			0.035  &  0.022  & 0.131   & 0.251   &  0.898    & 0.107\\
			\hline
			
			\hline
		\end{tabular}
	}
		\label{tabel:inference time}
		\vspace{-0.1in}
	\end{table}

	\subsection{Evaluation on Video Deblurring }
	Two synthetic datasets GOPRO and REDS are employed for quantitative and qualitative evaluation as well as computational efficiency, and one real-world dataset BSD is adopted for evaluating generalization ability. 
	
	\subsubsection{Comparison on Synthetic Datasets}
	When synthesizing blurry frames in GOPRO and REDS, more frames are averaged in REDS, thus resulting in more severe blur in REDS than in GOPRO.

	\emph{\textbf{Evaluation on GOPRO Dataset.}}
	% On GoPro dataset, we evaluate the deblurring performance on both blurry frames (Table \ref{tabel:gopro-sin}) and whole video frames (Table \ref{tabel:gopro-vi}). 
	%
	We retrain all these competing methods on the training set of GOPRO for a fair comparison. 
%	For Transformer-based method VRT, we finetune the model provided by authors on our dataset. It is worth noting that VRT is trained on Tesla A100 GPUs, but we do not have such large computing resources. Hence, we use a smaller patch size for finetuning.
	%
	Table \ref{tabel:gopro-vi} reports the comparison of PSNR and SSIM metrics for the competing methods. 
	One can see that our method outperforms CNN-based and RNN-based methods by a large margin, among which D$^2$Nets~\citep{shang2021bringing} also leverages sharp frames similar to our method. 
	Benefiting from {nearest} sharp features, our method and D$^2$Nets are generally better than other CNN-based methods. 
	However, in D$^2$Nets, optical flow is adopted to perform spatial alignment, possibly yielding texture loss and deformation artifacts. 
	And thus, our method is still much better than D$^2$Nets. 

	In comparison to Transformer-based method VRT and CDVD-TSPNL, the attention in temporal dimension can implicitly benefit from the {nearest} sharp features, and thus they are notably better than CNN-based methods. 
	Our method still achieves about $\textgreater$0.3dB PSNR gain than VRT. 
	Meanwhile, our method is much more efficient than VRT, as shown in Table \ref{tabel:inference time}, although their parameters are similar. 
	Considering the quantitative performance in Table \ref{tabel:gopro-vi} and computational cost in Table \ref{tabel:inference time}, our method has a good tradeoff between computational cost and performance.
	In terms of visual quality comparison in Fig. \ref{fig:gopro}, flow-based method, \eg, CDVD-TSPNL and RNN-based methods, \eg, STFAN and ESTRNN, all fail in recovering sharp texture. 
	MIMOUNet++ is an image deblurring method, and it has a better deblurring result, but it produces some undesirable artifacts, referring to yellow box in Fig. \ref{fig:gopro}.
	D$^2$Nets utilizes sharp frames by using optical flow for warping sharp frames to current frame, and then directly concatenating them with current frame. However, direct concatenation can not fully extract sharp textures for reconstructing current blurry frame.
	Our method can achieve sharper texture details by utilizing useful sharp information searched from detected sharp frames by global Transformer in the multi-scale scheme, and the boundary outline of the car is quite clearer than the results by the competing methods in Fig. \ref{fig:gopro}.

	\emph{\textbf{Evaluation on REDS Dataset.}}
	All methods are retrained on the REDS dataset, and the quantitative results are reported in Table \ref{tabel:reds-vi}. 
	Benefiting from attention mechanism, CDVD-TSPNL, VRT and our method are much better than the other methods. 
	%,
	Among them, our method achieves the best performance in comparison with competing methods in terms of both PSNR and SSIM.  
	Fig. \ref{fig:reds} shows the visual quality comparison, from which one can see that our method can recover sharper texture details, due to the guidance of {nearest} sharp features, 
	while the results by other methods still suffer from mild blur or over-smoothing textures. 
	From Fig. \ref{fig:reds}, one can see unsatisfactory distortion of the street lamp post in the red box of VRT~\citep{liang2024vrt} and the details of fences in yellow box are not clear. 
		
	\begin{figure*}[!t]\footnotesize
		\centering
		\setlength{\tabcolsep}{0pt}
		
		\begin{tabular}{cclcclcclcclcclccl} %{ccccc}
			\includegraphics[width=0.99\linewidth]{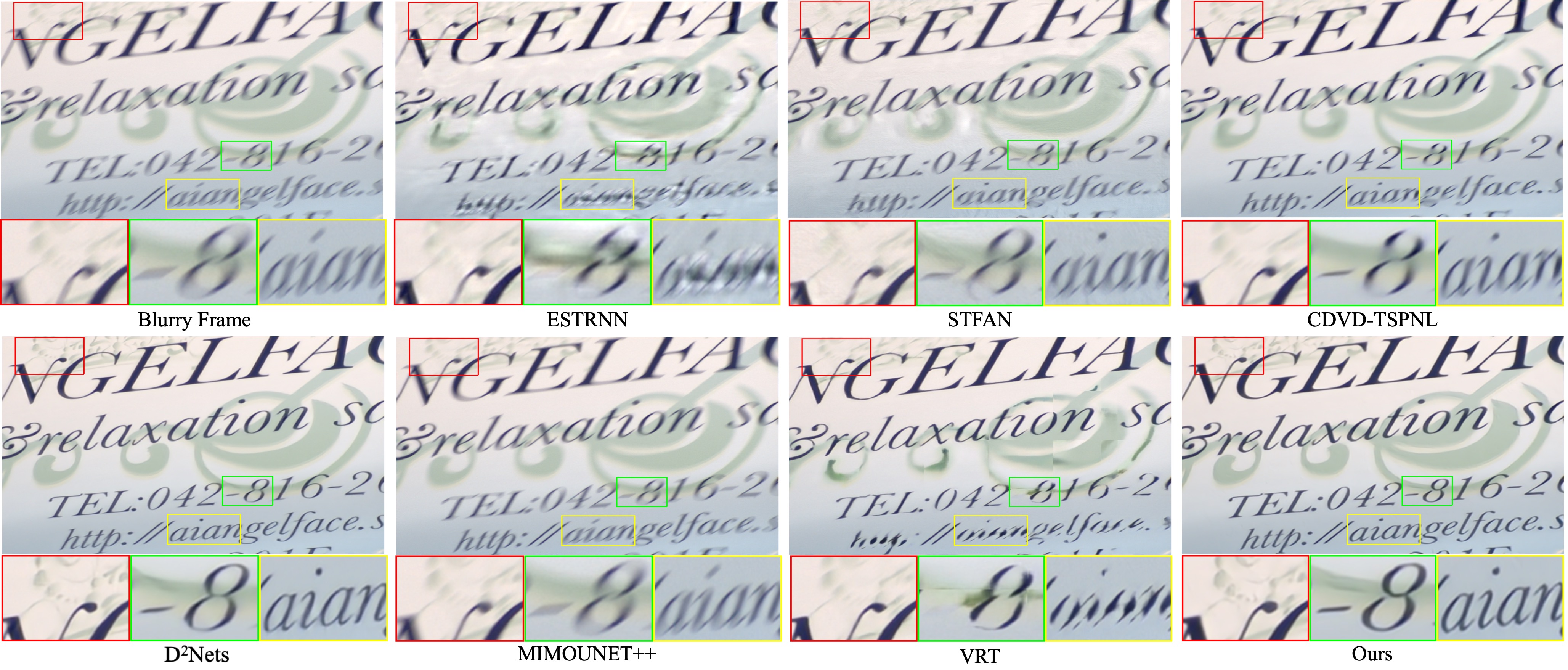}\\
			
		\end{tabular}
		\vspace{-0.8em}
		\caption{Visual comparison of deblurring results on real-world blurry video.  
		}
		\label{fig:bsd}
	\end{figure*}	
	
	\subsubsection{Generalization Evaluation on Real-world BSD Dataset}
	We evaluate generalization ability of these methods on real-world blurry videos, where the models of competing methods trained on REDS dataset are applied to handle real-world blurry videos from BSD dataset. 
	For the comparison, we adopt two settings. 
	First, we use our blur-aware detector to find sharp frames in BSD testing dataset, and we select videos that contain more than 40$\%$ sharp frames as real non-consecutively blurry videos to satisfy the condition of {nearest} sharp features.
	We obtain 13 non-consecutively blurry videos from BSD dataset for evaluating generalization ability, and we denote it as BSD$^-$. 
	The results are reported in Table \ref{tabel: bsd}. 
	Second, we adopt all the testing videos from BSD for evaluation, and the results are reported in Table \ref{tabel: bsd_all}. 
	In this setting, there are videos without sharp frames, \ie, $\bm{G}_i^+$ and $\bm{G}_i^-$ cannot be found by detector, and we use $\bm{B}_{i-2}$ and $\bm{B}_{i+2}$ to substitute $\bm{G}_{i}^-$ and $\bm{G}_{i}^+$, respectively.

	As shown in Table \ref{tabel: bsd}, our method achieves the highest PSNR and SSIM values, indicating better generalization ability of our method. 
	We note that VRT is much inferior to our method, although it achieves comparable metrics on GOPRO dataset in Table \ref{tabel:gopro-vi} and REDS dataset in Table \ref{tabel:reds-vi}. 
	This may be attributed that VRT is overfitted to the training set, resulting in poor generalization ability. 
	In Table \ref{tabel: bsd_all}, the results for different capturing settings are reported, where our method achieves the highest PSNR and SSIM metrics for all the three settings. 
	We note that our method still works well for real-world videos without sharp frames, making our method applicable in practical applications. 
	Also Transformer-based method VRT has the worst generalization ability. 
	As shown in Fig. \ref{fig:bsd}, our method achieves the most visually plausible deblurring results with sharper textures, while STFAN, VRT, ESTRNN and MIMOUNet++ cannot fully remove severe blur. 
	CDVD-TSPNL and D$^2$Nets generate undesirable artifacts on letters in red box. 
	\begin{table*}[!tb]\small
		\setlength{\tabcolsep}{5pt}
		\caption{Quantitative comparison of generalization ability on BSD$^-$ dataset. 
		}
		\centering
		\resizebox{\columnwidth}{!}{
		\begin{tabular}{c|ccccccc|c}
			\hline
			
			\hline
			Method &  DMPHN &  STFAN & 
			% CDVD-TSP\citep{pan2020cascaded}  &
			ESTRNN  &  MIMO-UNet++  & D$^2$Nets &CDVD-TSPNL& VRT    & Ours  \\
			\hline
			PSNR &  30.39  & 27.69  & 
			% 30.40     &
			25.984    &   28.77  &  {30.58}  & 28.79 &  26.73 &  \textbf{32.10}  \\
			SSIM &   0.924  &  0.860 &   
			% 0.926    &
			0.843    &  0.884  & {0.933} &0.900 & 0.870  &  \textbf{0.937}  \\
			
			\hline
			
			\hline
		\end{tabular}
	}
		\label{tabel: bsd}
		\vspace{-0.1in}
	\end{table*}
	
	\begin{table}[!tb]\small
		\setlength{\tabcolsep}{4pt}
		\caption{Quantitative comparison of generalization ability on the whole BSD dataset. 
		}
		% \centering
		\resizebox{\columnwidth}{!}{
		\begin{tabular}{c|ccccccc|c}
			\hline
			
			\hline
			Method & DMPHN &  STFAN & 
			ESTRNN  &  MIMO-UNet++  & D$^2$Nets &CDVD-TSPNL& VRT    & Ours  \\
			\hline
			1ms-8ms  &  29.94/0.905  &  27.49/0.845 & 
			26.32/0.827   &  29.27/0.874  & 30.14/0.911 &28.16/0.870& 25.99/0.845    & \textbf{31.40/0.918}  \\
			2ms-16ms  & 28.47/0.885 &  25.71/0.799 & 
			24.36/0.779   &  26.98/0.830   & 28.55/0.888 &26.37/0.835& 23.67/0.785   &  \textbf{29.93/0.895} \\
			3ms-24ms   &   27.64/0.877 &  26.17/0.825 & 
			25.29/0.803  &  27.40/0.850  & 28.40/0.891 &26.78/0.851& 25.07/0.822    & \textbf{29.25/0.891}  \\
			\hline
			
			\hline
		\end{tabular}
	}
		\label{tabel: bsd_all}
		\vspace{-0.1in}
	\end{table}

	\subsection{Evaluation on Event-Driven Video Deblurring} \label{sec:ext}
		
	\begin{figure}[!t]\footnotesize
		\centering
		\setlength{\tabcolsep}{0pt}
		
		\begin{tabular}{cclcclcclcclcclccl}
			\includegraphics[width=.99\linewidth]{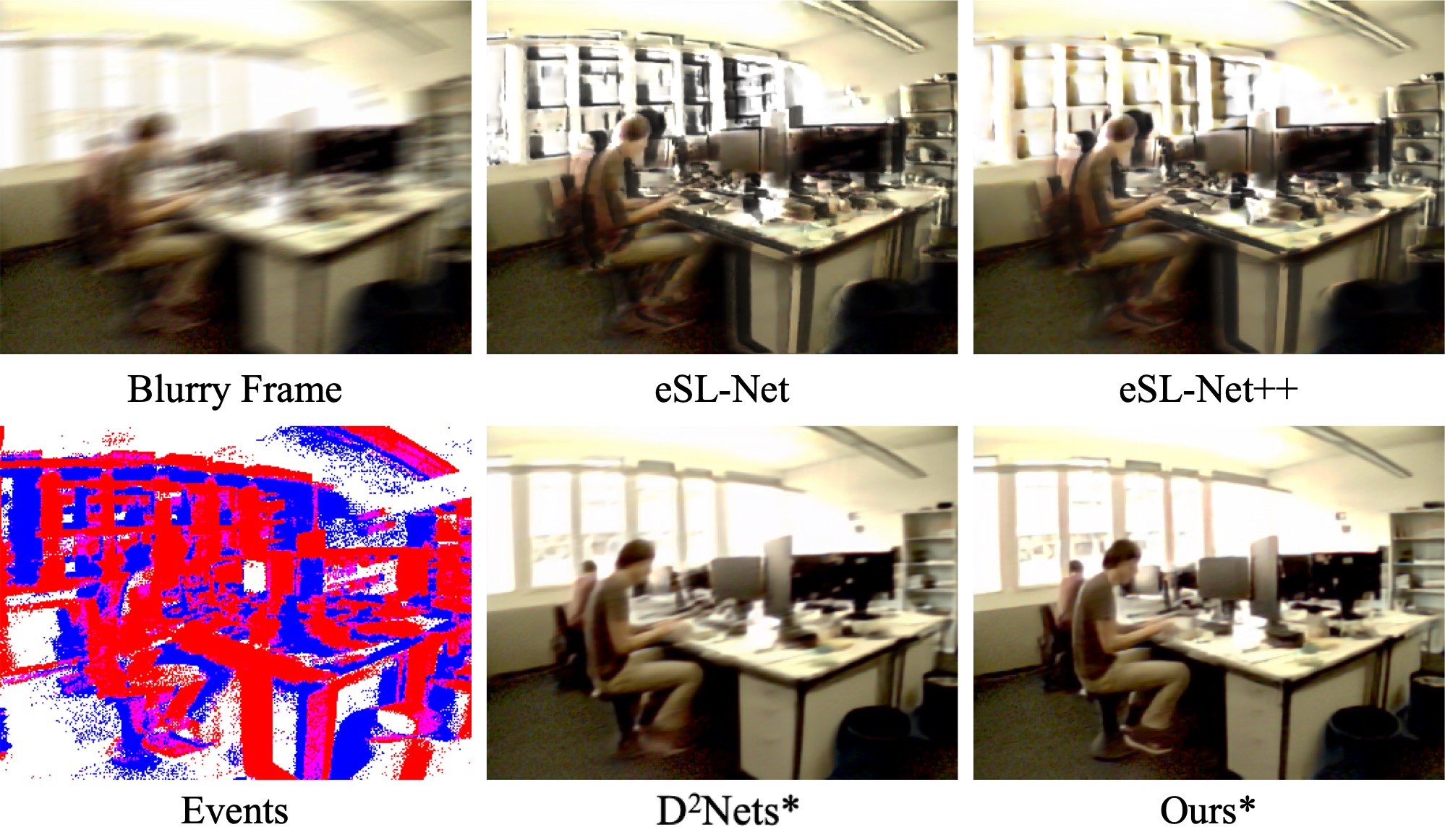}	
			
		\end{tabular}
		\vspace{-0.8em}
		\caption{Visual comparison of deblurring results on CED dataset. Please zoom in for better view.}
		\label{fig:event}
	\end{figure}	
	Our method is trained on CED dataset, dubbed Ours*. 
	As for the competing methods, D$^2$Nets* is also re-trained on CED dataset, while the models of eSL-Net~\citep{wang2020event} and eSL-Net++~\citep{yu2022learning} provided by the authors are directly adopted for evaluation since their training codes are not available. 
	As shown in Table \ref{tabel:event}, one can see our method can achieve higher PSNR and SSIM values than the other methods. 
	eSL-Net~\citep{wang2020event} is very dependent on the number of utilized events. More events will lead to darkness, and fewer events will lead to not removing blur. 
	From Fig. \ref{fig:event}, event-based method eSL-Net~\citep{wang2020event} generated unpleasurable darkness noise. The results by eSL-Net++~\citep{yu2022learning} still suffer from  slight blur. And D$^2$Nets*~\citep{shang2021bringing} can remove much blur but the restored details are limited due to the simple utilization of adjacent frames and events. The hands and shoes are blurry in the results of D$^2$Nets* in Fig. \ref{fig:event}. One can see our method can achieve better results compared with eSL-Net++ and D$^2$Nets*.
	
\begin{figure}[!t]\footnotesize
		\centering
		\setlength{\tabcolsep}{0pt}
		
		\begin{tabular}{cclcclcclcclcclccl}
			\includegraphics[width=.99\linewidth]{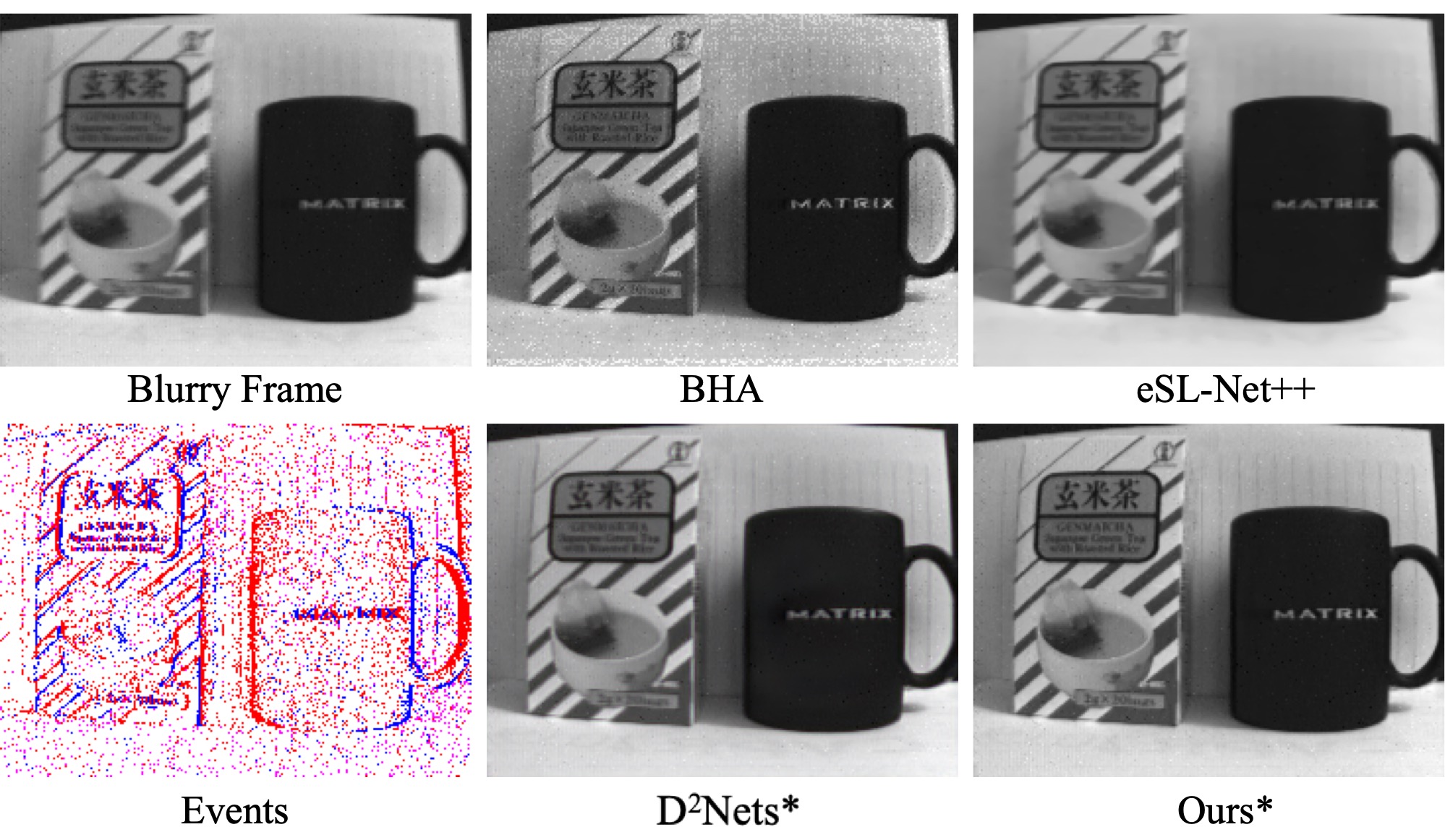}

		\end{tabular}
		\vspace{-0.8em}
		\caption{Visual comparison of deblurring results on RBE dataset. Please zoom in for better view.}
		\label{fig:event_real}
	\end{figure}

	In addition, we also compare our method with the other methods on RBE dataset~\citep{pan2019bringing}. 
	The models of D$^2$Nets*~\citep{shang2021bringing} and our method trained on CED dataset are adopted for testing. 
	For BHA~\citep{pan2019bringing} and eSL-Net++~\citep{yu2022learning}, we use the pre-trained models provided by authors, since they do not release training codes. 
	In Fig. \ref{fig:event_real}, one can see eSL-Net++ fail to deblur on real blurry frames. BHA and D$^2$Nets* can remove the blur but suffer from noises and darkness artifacts, while our method can achieve more visually plausible deblurring results compared with the other methods.

	\begin{table}[!tb]\small
		\caption{Quantitative comparison of deblurring results on CED dataset.}
		\centering
		\begin{tabular}{c|ccc|c}
			\hline
			
			\hline
			Method &  eSL-Net  & eSL-Net++  & D$^2$Nets*    & Ours*   \\
			\hline
			PSNR &23.12  & 25.797   & 31.72   &  \textbf{33.55}      \\
			SSIM   & 0.716   & 0.7545   &0.912  & \textbf{0.936}         \\
			
			\hline
			
			\hline
		\end{tabular}
		\label{tabel:event}
	\end{table}
	
	\subsection{Ablation Study}

	\subsubsection{Accuracy of Blur-aware Detector} %\label{sec:detector}
	
	Table \ref{tabel:detector} lists the accuracy of blur-aware detector on two datasets by using different training loss functions, where we utilize the detector trained on REDS to detect videos from GOPRO for evaluating generalization ability. 
	One can see that detector $\mathcal{F}_\text{Detector}$ trained with supervised contrastive loss $\mathcal{L}_\text{contra}$ has slightly lower accuracy than that using only $\mathcal{L}_\text{ce}$ on GOPRO dataset. 
	However, benefiting from contrastive learning, the features $\bm{z}$ between sharp and blurry frames can be forced as distinguishable as possible, leading to better generalization ability. 
	That is to say the detector trained by cross-entropy loss may be overfitted to training set while failing to handle real-world blurry videos. 
	From the third row in Table \ref{tabel:detector}, the detector trained with supervised contrastive loss performs much better.
	Moreover, we also detect sharp frames on real-world dataset BSD, where almost 80$\%$ of the detected sharp frames are consistent with human visual observation, further indicating the advantages of $\mathcal{L}_\text{contra}$ on generalization ability.
	In Table \ref{tabel: bsd_all}, for the videos without sharp frames as guidance, our method is still superior to the competing methods. 
		\begin{table}[!tb]\small
		\caption{ The accuracy of BiLSTM detector by using different loss. }
		\centering
		\begin{tabular}{c|cc}
			\hline
			
			\hline
			Loss  &  Only $\mathcal{L}_\text{ce}$ &  $\mathcal{L}_\text{ce}$ \& $\mathcal{L}_\text{contra}$  \\
			
			\hline
			GOPRO  &  99.9$\%$   &   98.4$\%$  \\
			%\citep{nah2017deep}
			% \hline
			REDS  &  87.2$\%$   &  87.2$\%$    \\
			%~\citep{nah2017deep}
			\hline
			Generalization  &  90.5$\%$   &  95.0$\%$    \\
			
			\hline
		\end{tabular}
		\label{tabel:detector}
	\end{table}

	\subsubsection{Effectiveness of Hybrid Transformer for Video Deblurring}
\begin{figure*}[!t]\footnotesize
	\centering
	\setlength{\tabcolsep}{1pt}
	\begin{tabular}{cclcclcclcclcclccl}
		\includegraphics[width=.99\textwidth]{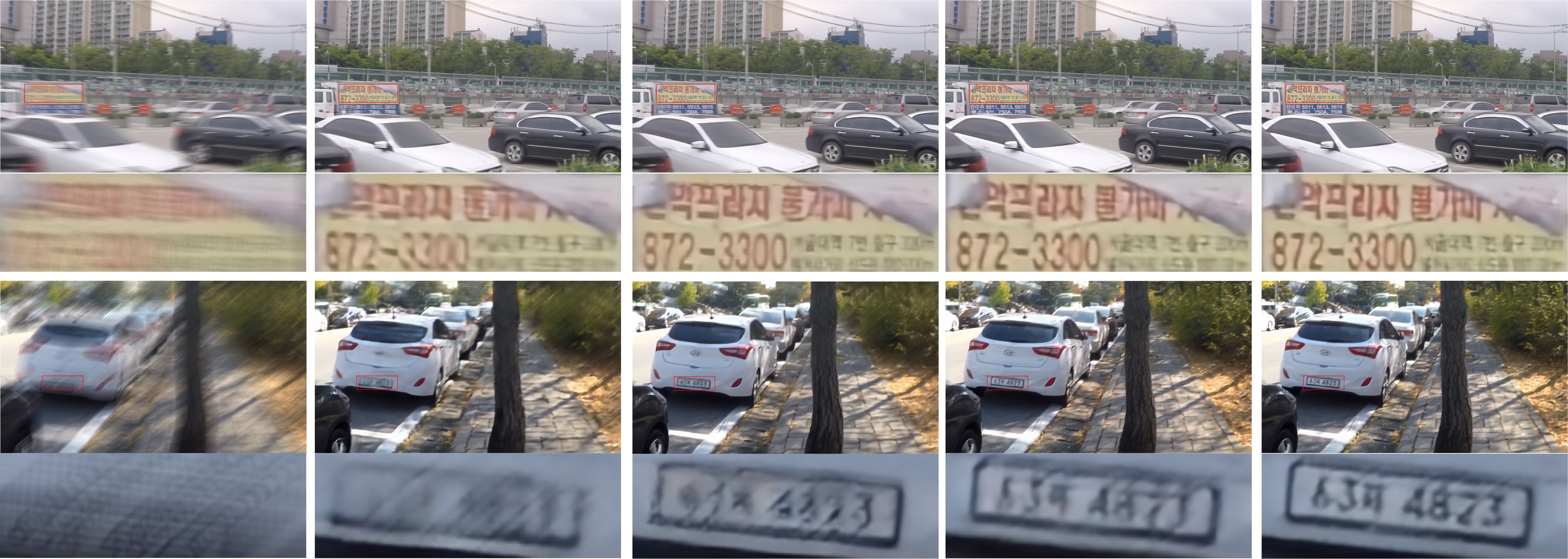}	
	\end{tabular}
	\caption{Visual comparison of component analysis on GOPRO dataset. 
		The first column is blurry frame, and 2 $ \sim $ 5 columns correspond to the results of 1 $ \sim $ 4 rows in Table \ref{tabel:ablation}.
		Zoom in for better view.}
	\label{fig:ablation}
\end{figure*}

	We evaluate the contribution of each component of proposed video deblurring framework. 
	The experiments are conducted on GOPRO dataset. 
	Self-attention means taking only current frame as input of $f_\text{cswt}$, and using self-attention in W-MCA and SW-MCA instead of cross-attention.
	And the variant with self-attention and global attention takes current blurry frame and its {nearest} sharp frames as input, while discarding its two neighboring frames
	% But it does not utilize adjacent frames.
	From Table \ref{tabel:ablation} and Fig. \ref{fig:ablation}, one can see that {nearest} sharp features aggregated by global attention bring about 1dB PSNR gain, indicating the effectiveness of global Transformer for aggregating {nearest} sharp features. 
	By adopting cross-attention, features from neighboring frames can be well exploited, leading to better performance. 
	In our final model, the superior results validated the necessity of adopting cross-attention for aggregating neighboring frames and global attention for aggregating {nearest} sharp frames for the best deblurring performance. 

	Besides these key variants of proposed framework in Table \ref{tabel:ablation}, we further discuss another case of cross-attention. 
	Since it is a natural strategy to directly take detected sharp frames and neighboring frames as input of local Transformer, we directly take five frames $\{\bm{G}_i^-,\bm{B}_{i-1}, \bm{B}_i, \bm{B}_{i+1}, \bm{G}_i^+\}$ as input of $f_\text{cswt}$ as input while removing global Transformer. 
	In this case, we obtain the results with PSNR 34.19dB and SSIM 0.952, which is much inferior to our final model. 
	The reason can be attributed from two aspects: 
	(i) Neighboring frames and detected sharp frames have different temporal distances, bringing more difficulty for the feature fusion by using one ${f}_\text{cswt}$. 
	% for feature fusion is difficult for learning due to different temporal distance, and 
	(ii) Detected sharp frames from a long temporal distance may have dramatic scene changes with the blurry frame, which may be beyond the range of window-based cross-attention.

	\begin{table}[!tb]\small
	\caption{ Component analysis on the GOPRO dataset. 
	}
	\centering
	\begin{tabular}{c|c|c|ccc}
		\hline
		
		\hline	
		Self-attention  &   Cross-attention  &   Global attention & PSNR  & SSIM \\
		\hline
		\Checkmark &  \XSolidBrush  &   \XSolidBrush    &  35.28             &   0.938    \\
		\Checkmark &   \XSolidBrush  &    \Checkmark      &   36.24   &  0.953      \\
		\XSolidBrush &  \Checkmark   &    \XSolidBrush    &  36.94    &  0.959      \\
		\XSolidBrush &  \Checkmark    &  \Checkmark   &  37.33    & 0.962       \\
		
		\hline
		
		\hline
	\end{tabular}
	\label{tabel:ablation}
	\vspace{-0.1in}
\end{table}

	\subsubsection{Effectiveness of Nearest Sharp Frames} \label{sec:nsf}
\begin{figure}[!t] \footnotesize
		\begin{tabular}{lcccccc}
			\ \ \
			\includegraphics[width=0.9\linewidth]{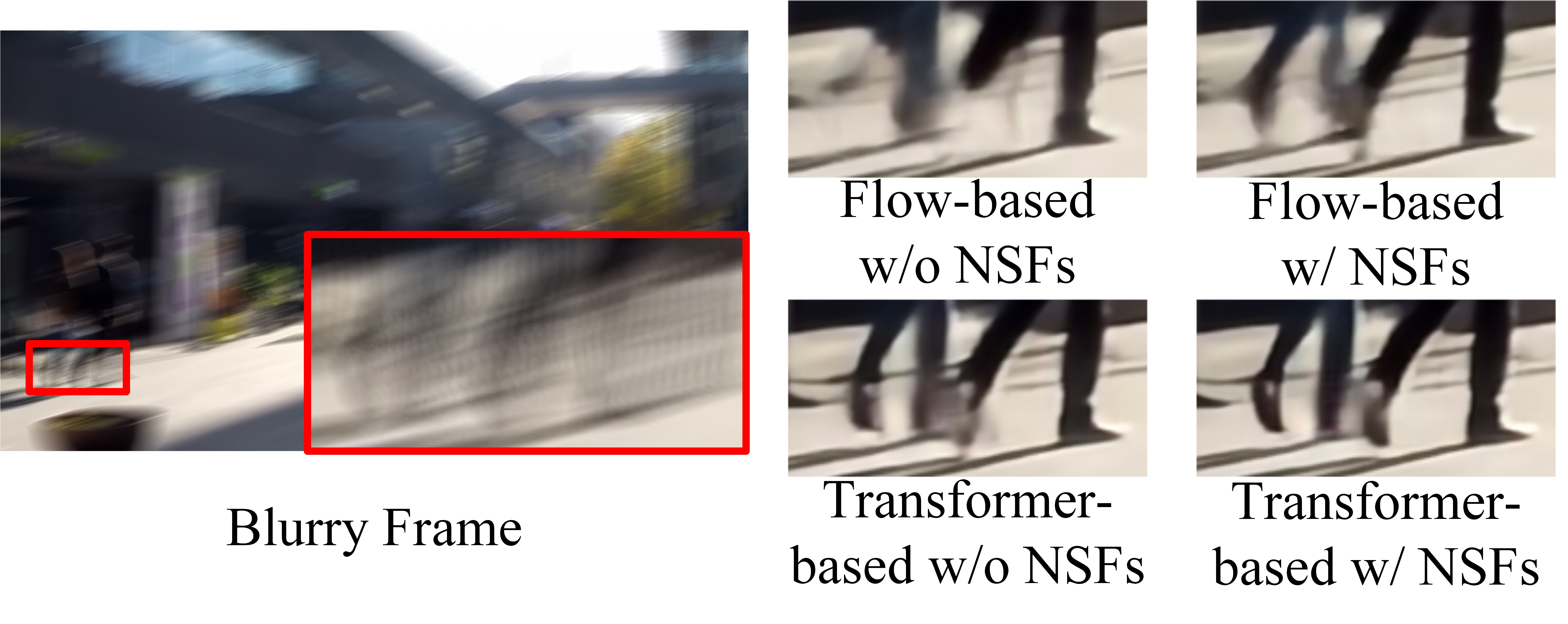} 
			\ \ \
		\end{tabular}
	\vspace{-1em}
		\caption{{Effect of NSFs on non-consecutively blurry video deblurring.}}
		\label{fig:nsf}
	\end{figure}

	In order to verify Nearest Sharp Frames (NSFs) are indeed helpful for deblurring on non-consecutively blurry frames. We conduct experiments on different algorithms with or without NSFs. We denote the algorithm using optical flow for alignment as flow-based, and it warps other frames to current frame and then feeds them into a U-Net for reconstructing. 
	For fair comparison, we set the parameters of the U-Net to be similar to ours.
	And we denote our algorithm using transformer for long-range dependency modelling as transformer-based. The result can be seen in Table \ref{tabel:nsf}. 
	One can see algorithms with NSFs can achieve higher PSNR/SSIM and better visual results, and transformer-based algorithm with NSFs can achieve most pleasuring results compared with others in Fig. \ref{fig:nsf}. Sharp texture information in NSFs can be utilized for guiding current frame restoration in both flow-based and transformer-based methods.
	\begin{table}[!tb]\small
		\setlength{\tabcolsep}{15pt}
		\caption{ Effectiveness of NSFs in different algorithms on GOPRO.}
		\centering
		\begin{tabular}{c|c|c}
				\hline
				
				\hline
				Input  & w/o NSFs & w/ NSFs  \\
				
				\hline
%				Flow-based &  33.75/0.925   &   33.84/0.931   \\
				Flow-based &  34.75/0.935   &   35.18/0.943   \\
				
				\hline
				Transformer-based &  36.94/0.959   &  37.33/0.962     \\

				\hline
			\end{tabular}
		\label{tabel:nsf}
	\end{table}

\begin{figure*}[!htb]\footnotesize
		\centering
		\setlength{\tabcolsep}{0pt}
		
		\begin{tabular}{c}
			
			\includegraphics[width=\textwidth]{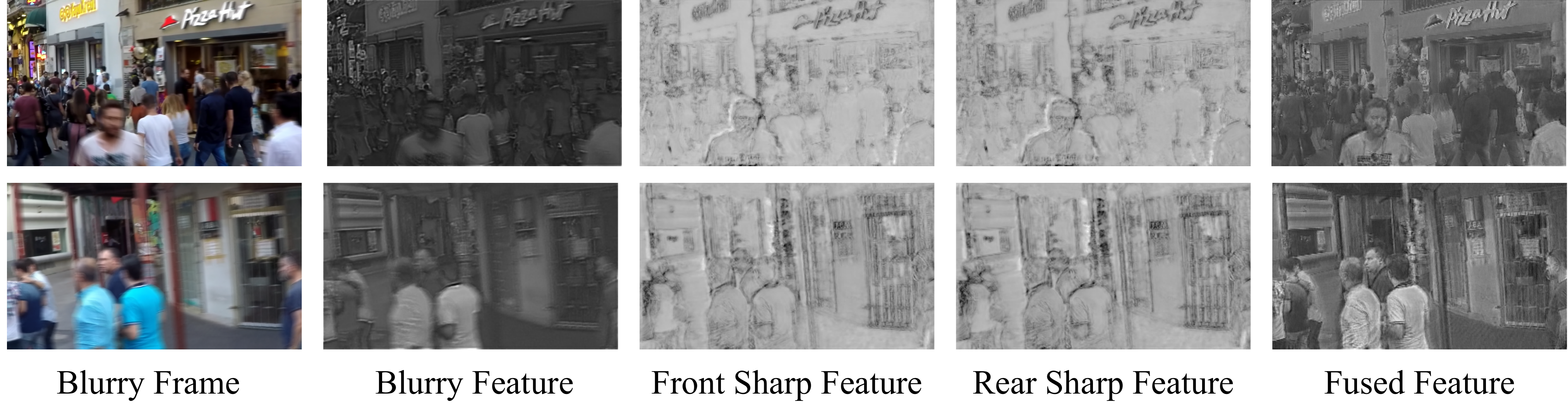}\\
			
		\end{tabular}
		\caption{Visualization of nearest sharp features. }
		\label{fig:visual}
	\end{figure*}

	{Then we take the third scale as an example to visualize the blurry features extracted from blurry frames, and also visualize corresponding front and rear sharp features obtained using Global Transformer. From Fig.~\ref{fig:visual}, one can see that the long-term sharp features obtained through our Global Transformer have clear texture, greatly promoting the deblurring effect of the method. Regarding the searching range $N$ for sharp frames, we supplemented the ablation study in Table~\ref{table:search} by gradually expanding $N$. As the search range $N$ increases, the performance gradually improves. When $N$ is greater than 7, the performance tends to stabilize or slightly decrease, indicating that some sharp frames are far away from the current frame, and blurry objects move out of the frame over time, which is not very helpful for restoring the current blurry frame. }
\begin{table*}[!htb]\small
	\setlength{\tabcolsep}{6pt}
	\caption{Impact of Searching range $N$ on REDS dataset. 
	}
	\centering
	\begin{tabular}{c|ccccc}
		\hline
		
		\hline
		$N$ & 3  & 5 & 7  & 10  &  15  \\
		\hline
		PSNR     & 40.81   & 41.31     & 41.45       &  41.43  & 41.40 \\
		SSIM     & 0.975    & 0.980   & 0.982     & 0.981  &  0.981\\
		
		\hline
		
		\hline
	\end{tabular}
	\label{table:search}
\end{table*}
		
		\section{Conclusion}\label{sec:conclusion}
		{In this paper, we introduce a novel video deblurring framework designed to aggregate the nearest sharp features. Our approach commences with the training of a blur-aware detector, which is pivotal in discerning between blurry and sharp frames. This capability is instrumental in pinpointing the nearest sharp frames, thereby providing a reference for the deblurring process.
			Subsequently, we employ a hybrid Transformer for video deblurring. It involves the utilization of window-based local Transformers to aggregate features from adjacent frames and global Transformers to integrate features from the nearest sharp frames within a multi-scale framework. The integration of an event fusion module further enables the extension of our method to event-driven video deblurring.
			Our extensive experimental evaluations on benchmark datasets substantiate the efficacy of our approach, as evidenced by superior performance in quantitative metrics and visual quality. The blur-aware detector and the video deblurring network we have developed demonstrate enhanced generalization capabilities, particularly when dealing with real-world blurry videos.
			In future work, we envision the potential expansion of our proposed method into other pertinent domains, such as video super-resolution, joint video deblurring and multi-frame interpolation, further broadening the applicability of our research.}
		
	\section*{Acknowledgements}
	This work was supported in part by the National Natural Science Foundation of China (62172127 and U22B2035), the Natural Science Foundation of Heilongjiang Province (YQ2022F004).
%\appendix
%\section{Example Appendix Section}
%\label{app1}
%%% For citations use: 
%%%       \cite{<label>} ==> [1]
%
%%%
%Example citation, See \cite{lamport94}.
%
%%% If you have bib database file and want bibtex to generate the
%%% bibitems, please use
%%%
%%%  \bibliographystyle{elsarticle-num} 
%%%  \bibliography{<your bibdatabase>}
%
%%% else use the following coding to input the bibitems directly in the
%%% TeX file.
%
%%% Refer following link for more details about bibliography and citations.
%%% https://en.wikibooks.org/wiki/LaTeX/Bibliography_Management

%\begin{thebibliography}{00}
%
%%% For numbered reference style
%%% Text of bibliographic item
%
%\bibitem{lamport94}
%  \textit{\LaTeX: a document preparation system},
%  Addison Wesley, Massachusetts,
%\end{thebibliography}
\bibliographystyle{elsarticle-num}  %# 这里不用改，对应的是elsarticle-num.bst文件
\bibliography{sn-bibliography}  %# 填写.bib文件的文件名

\end{document}